\documentclass{article}

\PassOptionsToPackage{numbers,sort&compress}{natbib}
\usepackage[final,main]{neurips_2026}

\usepackage[utf8]{inputenc}
\usepackage[T1]{fontenc}
\usepackage{microtype}
\usepackage{graphicx}
\usepackage{subcaption}
\usepackage{booktabs} 
\usepackage{multirow} 
\usepackage{pifont}   
\usepackage{hyperref}
\usepackage{latexml}
\iflatexml
  \usepackage{authblk}
\fi
\usepackage{url}
\usepackage{xcolor}
\usepackage{algorithm}
\usepackage{algorithmic}

\usepackage{float}     
\usepackage{amsmath}
\usepackage{amssymb}
\usepackage{mathtools}
\usepackage{amsthm}
\usepackage{tabularx}
\usepackage{makecell}
\usepackage{tikz}
\usepackage{pgfplots}
\usepackage{pgfplotstable}
\pgfplotsset{compat=1.18}
\usepgfplotslibrary{groupplots,fillbetween}

\usepackage[textsize=tiny]{todonotes}

\usepackage[capitalize,noabbrev]{cleveref}

\theoremstyle{plain}

\theoremstyle{definition}

\theoremstyle{remark}

\title{TENET: Time-point Encoding Network for Multivariate Time Series Anomaly Detection}

\iflatexml
\author{\textbf{Dongchan Cho}\textsuperscript{*}}
\affil{Industrial AI Lab, SimPlatform Co. Ltd.\\
  Affiliate Research Institute\\
  \href{mailto:cdc0213@simplatform.com}{\texttt{cdc0213@simplatform.com}}\\
  \textsuperscript{*}Corresponding author.}
\author{\textbf{Juwon Hwang}}
\affil{Industrial AI Lab, SimPlatform Co. Ltd.\\
  Affiliate Research Institute\\
  \href{mailto:jwhwang@simplatform.com}{\texttt{jwhwang@simplatform.com}}}
\author{\textbf{Keumyeong Kang}}
\affil{Industrial AI Lab, SimPlatform Co. Ltd.\\
  Affiliate Research Institute\\
  \href{mailto:kykang@simplatform.com}{\texttt{kykang@simplatform.com}}}
\author{\textbf{Jiho Han}}
\affil{Industrial AI Lab, SimPlatform Co. Ltd.\\
  Affiliate Research Institute\\
  \href{mailto:jihohan@simplatform.com}{\texttt{jihohan@simplatform.com}}}
\author{\textbf{Namsoon Jung}}
\affil{Industrial AI Lab, SimPlatform Co. Ltd.\\
  Affiliate Research Institute\\
  \href{mailto:nsj@simplatform.com}{\texttt{nsj@simplatform.com}}}
\else
\author{%
  Dongchan Cho\thanks{Corresponding author.}
  \quad
  Juwon Hwang
  \quad
  Keumyeong Kang
  \quad
  Jiho Han
  \quad
  Namsoon Jung \\
  Industrial AI Lab, SimPlatform Co. Ltd. Affiliate Research Institute \\
  \texttt{\{cdc0213, jwhwang, kykang, jihohan, nsj\}@simplatform.com}
}
\fi

\hypersetup{pdfauthor={Dongchan Cho, Juwon Hwang, Keumyeong Kang, Jiho Han, Namsoon Jung}}

\begin{document}

\maketitle

\begin{abstract}
In unsupervised multivariate time series anomaly detection (MTSAD), local variation can inflate reconstruction and forecasting errors, making observation-space discrepancy an unreliable indicator of abnormal system behavior.
Since latent-space scoring can mitigate this limitation, we reframe one-class MTSAD as current-point system-state compatibility estimation and propose \textbf{TENET}, the \textbf{T}ime-point \textbf{E}ncoding \textbf{Net}work for Multivariate Time Series Anomaly Detection.
We define the current-point system state as a latent representation of the target time point, constructed from variable-level temporal context in the input window.
TENET constructs this state through a temporal layer that retrieves window context relative to the current point, encodes time with current-point-centered embeddings, and adaptively mixes the retrieved context with the current representation.
With this design, TENET treats the input window not as the object to reconstruct, forecast, summarize, or score, but as context for constructing a latent representation evaluated with a fixed standard-normal negative log-likelihood score.
Across benchmark datasets, TENET achieves state-of-the-art performance.
Additional experiments further show that TENET remains effective under alternative one-class objectives and outperforms substitutions based on generic sequence encoders, suggesting that its advantage comes from decision-aligned representation construction rather than from a specialized scoring rule.
These results indicate that one-class MTSAD can be strong when the latent representation is designed around the current time point.
\end{abstract}

\section{Introduction}
\label{sec:introduction}

Detecting anomalies in multivariate time series requires reasoning about the nonlinear dependencies that characterize normal system behavior~\cite{ACM-Survey-2024}. Because real-world anomalies constitute an open set and are often rare or previously unseen~\cite{Deep_AD_Survey-2021, Unsupervised_Model_Selection-2022, QuoVadisTAD-2024, TSAD_comprehensive_eval-2022}, supervised detection is often impractical.
As a result, multivariate time series anomaly detection (MTSAD) has largely shifted toward unsupervised learning, where the detector must infer normality from normal operation alone~\cite{OmniAnomaly-2019, SARAD-2024, CrossAD-2025, CATCH-2025, OracleAD-2025}.
Observation-space discrepancy has therefore become a natural and often effective anomaly signal in unsupervised MTSAD, as reflected in methods based on reconstruction residuals~\cite{AD_AE_NDR-2014, OmniAnomaly-2019, TranAD-2022, iTransformer-2023, ModernTCN-2023, SARAD-2024}, forecasting errors~\cite{TTMs-2024, Timer-2024, Chronos-2024, OracleAD-2025}, or local temporal inconsistencies~\cite{ScatterAD-2025}.

However, using observation-space discrepancy as an anomaly score relies on a strong surrogate assumption~\cite{NNContribute-2022, TSAD-illusion-2021}. Reconstruction and forecasting errors are meaningful only when the observation-space target can be reproduced or predicted with sufficient fidelity and when the remaining mismatch mainly reflects abnormal system behavior. This assumption can be fragile in real-world multivariate data, where normal signals may contain local variation, high-frequency fluctuations, or discontinuous changes that are difficult to reconstruct or predict even when the system remains normal. In such cases, a large observation-space error may reflect modeling error in the reconstruction or prediction task rather than an abnormal state. This limitation motivates latent one-class scoring, where abnormality is assessed through compatibility with a normal state space rather than through raw mismatch alone.

While one-class MTSAD addresses this surrogate problem by scoring latent normality, its semantic advantage has not been fully exploited. Existing latent-space anomaly detectors have largely advanced by changing the objective used to shape or score the normal space, such as hypersphere or center compactness~\cite{DeepSVDD-2018, THOC-2020}, density or energy estimation~\cite{DAGMM-2018}, and probabilistic likelihood-style scoring~\cite{OmniAnomaly-2019}, while leaving the construction of the scored representation comparatively underexamined. This matters because a one-class objective can only evaluate the representation it is given. If that representation is obtained by generic sequence compression or window-level summarization, the detector may still be scoring an object that is misaligned with the actual decision target. 

In point-wise MTSAD, the target is the current time point, not the window itself. We use \emph{state} to denote the latent representation evaluated for normality at that target point. Because the input window contains both target-relevant and target-irrelevant history, a window-level summary can dilute the information needed for the current decision. The fundamental question is therefore not only how to score normality in latent space, but what state should be constructed before scoring. We argue that the window should be used as context for the current point, so the encoder actively retains history that helps explain this state rather than uniformly accumulating the past or preserving information for reconstruction and forecasting. At the multivariate level, this current-centered temporal context should be formed separately for each variable and then composed into a \emph{current-point system state} before scoring. This shifts the bottleneck in one-class MTSAD from designing stronger normality objectives to constructing a decision-aligned state that makes one-class scoring meaningful.

Motivated by this view, we reformulate one-class MTSAD as estimating the compatibility of a current-point system state with normal behavior and propose \textbf{T}ime-point \textbf{E}ncoding \textbf{Net}work for Multivariate Time Series Anomaly Detection (\textbf{TENET}). TENET is a current-point-centric encoder that explicitly constructs the current-point system state before compatibility assessment. Its temporal layer uses the current point as an anchor to retrieve relevant recent history for each variable, after which inter-variable dependencies are modeled over the resulting current-point state tokens and aggregated into the current-point system state. This compatibility is then assessed during both training and inference, instantiated in our model as a fixed standard-normal NLL compatibility score.

Our main contributions are as follows:
\vspace{-0.5em}
\begin{itemize}
    \item We reformulate unsupervised one-class MTSAD as current-point system-state compatibility estimation and identify decision-aligned current-point system-state construction, rather than increasingly complex raw-space discrepancy modeling, as the central challenge.
    \item We propose TENET, a current-point-centric encoder that constructs the current-point system state by retrieving temporally relevant history with respect to the present and composing inter-variable dependencies before compatibility assessment.
    \item We show that TENET achieves state-of-the-art results across standard MTSAD benchmarks, while representation diagnostics indicate that its gains come from current-point-centric state construction rather than from the scoring objective alone.
\end{itemize}
\vspace{-0.5em}

\section{Related Work}
\label{sec:related_works}

\subsection{Latent-Space One-Class MTSAD}

Alongside observation-space discrepancy-based MTSAD, latent-space anomaly detection provides a complementary route that scores learned representations with one-class or probabilistic objectives instead of measuring reconstruction or prediction residuals.
Representative examples include Deep SVDD~\cite{DeepSVDD-2018}, THOC~\cite{THOC-2020}, DAGMM~\cite{DAGMM-2018}, and probabilistic latent-variable models such as OmniAnomaly~\cite{OmniAnomaly-2019}.
Despite substantial differences in formulation, these methods share the view that abnormality can be assessed in a learned normal state space rather than directly in the observation space.
Accordingly, much of the methodological variation in this line of work has centered on how the latent space is shaped, for example through hypersphere or center-based compactness~\cite{DeepSVDD-2018, THOC-2020}, Gaussian mixture density estimation in a compressed latent space~\cite{DAGMM-2018}, and variational latent-variable modeling~\cite{OmniAnomaly-2019}.
In comparison, our work focuses on the less explicit question of how to construct the latent state that is scored for current-point anomaly decisions in multivariate time series.

\subsection{Sequence Encoders for MTSAD}

Our work is also related to multivariate time series encoders used as backbones for forecasting, reconstruction, classification, and anomaly detection.
Common choices include recurrent encoders such as LSTM~\cite{LSTM-2014,EncDec-AD-2016}, convolutional and temporal-convolution models such as TCN~\cite{TCN-2018,ModernTCN-2023}, Transformer-style encoders~\cite{Transformer-2017,PatchTST-2022,Crossformer-2022,iTransformer-2023}, and more recent state-space models such as Mamba~\cite{Mamba-2024}.
Recent TSAD models also build on these backbones. xLSTMAD~\cite{xLSTMAD-2025} uses an xLSTM encoder--decoder for forecasting or reconstruction, while MemMambaAD~\cite{MemMambaAD-2025} combines a Mamba--TCN encoder with memory-guided reconstruction.
PaAno~\cite{PaAno-2026} instead learns local patch representations and scores their distances to a normal-patch memory bank.
These methods encode sequences or patches, whereas TENET uses the current point to query variable-level history before scoring the resulting representation.
Particularly relevant to our work are variable-aware designs that preserve variable-level representations before or during multivariate modeling.
PatchTST~\cite{PatchTST-2022} adopts channel-independent temporal encoding, while Crossformer~\cite{Crossformer-2022}, iTransformer~\cite{iTransformer-2023}, and OracleAD~\cite{OracleAD-2025} further introduce mechanisms for modeling dependencies across variable-level representations.
TENET differs by using the input window as variable-level context for constructing the state of the current time point, rather than compressing the entire window into a generic sequence representation.

\begin{figure*}[t]
    \centering
    \includegraphics[width=1\linewidth]{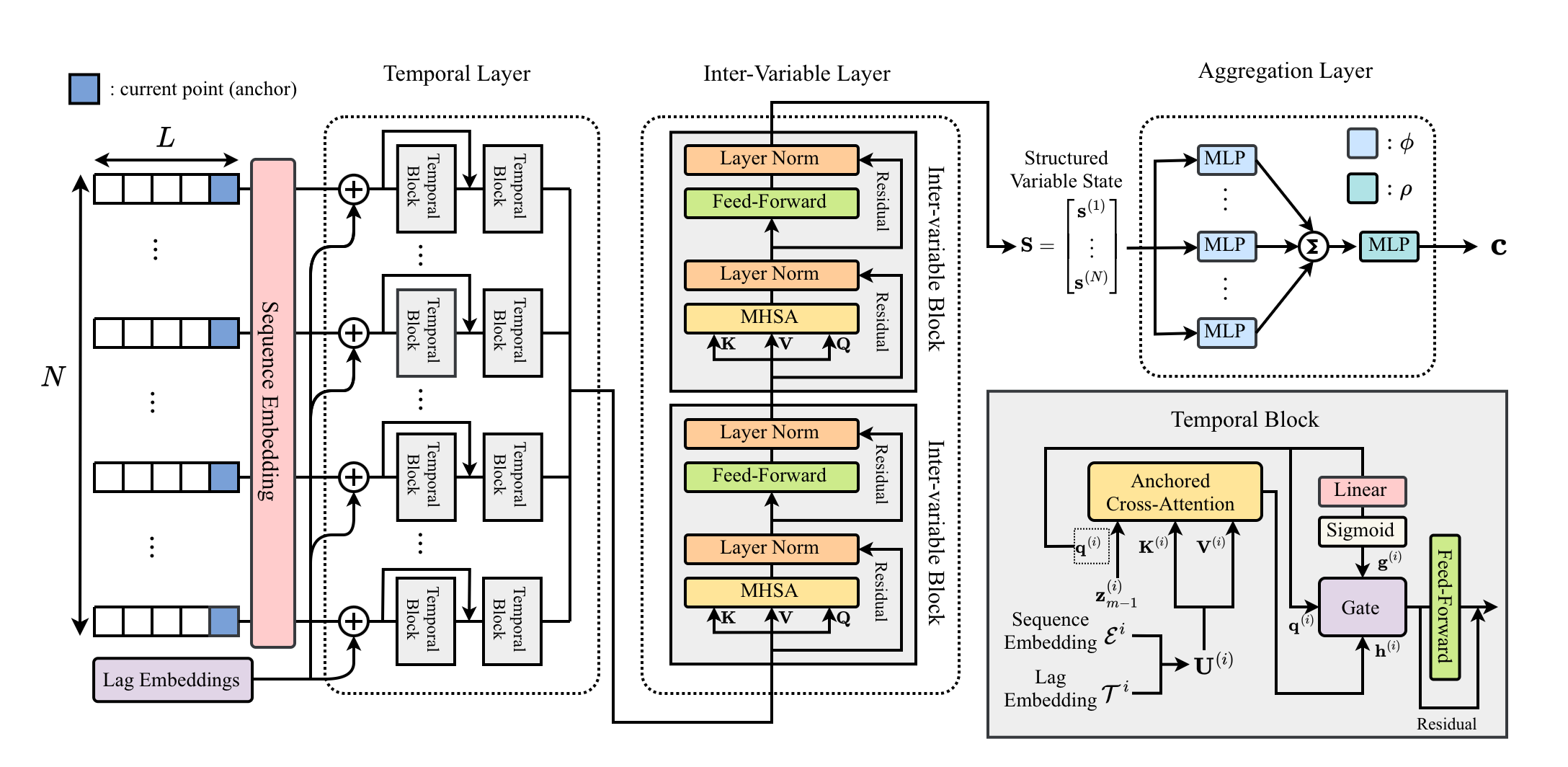}
    \vspace{-2em}
\caption{Architecture of TENET. A temporal layer, an inter-variable layer, and an aggregation layer construct the current-point system state from the input window. The inset shows the Temporal block, which uses \emph{anchored cross-attention} to retrieve history relative to the current point.}
    \label{fig:architecture}
    \vspace{-1em}
\end{figure*}

\section{Method}
\label{sec:method}
\vspace{-0.5em}
\subsection{Problem Setup}

We consider unsupervised multivariate time series anomaly detection (MTSAD).
Let the observed multivariate time series be denoted by $\mathcal{X}=\{\mathbf{x}_t\}_{t=1}^{T}$, where $\mathbf{x}_t \in \mathbb{R}^{N}$ is the observation vector at time $t$ over $N$ variables.
We assume that the training split contains only normal operation, whereas the test split may contain unknown anomalous events.
The detector operates on sliding windows.
For each time step $t \ge L$, we form an input window of length $L$ as
\begin{equation}
    \mathbf{X}_t = [\mathbf{x}_{t-L+1}, \dots, \mathbf{x}_t] \in \mathbb{R}^{L \times N}.
\end{equation}
Within this window, the last observation $\mathbf{x}_t$ is the current point to be evaluated, while the preceding sequence $\mathbf{x}_{t-L+1:t-1}$ provides local temporal evidence.
When the context is clear, we omit the subscript $t$ and write $\mathbf{X}$.
Our goal is to assign an anomaly score to each current point $\mathbf{x}_t$ based on the compatibility of its latent current-point system state with normal operation.
Here, the current-point system state denotes a learned representation for anomaly detection.
Rather than treating $\mathbf{X}_t$ itself as the decision object, we use it as context for constructing this state.

\vspace{-0.5em}
\subsection{Model Architecture}

TENET constructs the current-point system state $\mathbf{c}$ from a multivariate window $\mathbf{X}$ in three stages: a temporal layer $f_t$, an inter-variable layer $f_v$, and an aggregation layer $f_g$.
The first two stages produce a structured variable-level representation, while the last stage maps it to the current-point system state used for anomaly scoring.
Rather than encoding the entire multivariate window into a generic latent vector in a single step, this staged construction first forms current-point state tokens for individual variables, then composes them across variables, and only afterwards produces the current-point system state $\mathbf{c}$.
A detailed design rationale is provided in Appendix~\ref{appendix:state-geometry-rationale}.
\vspace{-0.5em}
\paragraph{Temporal Layer.}
To reduce cross-variable interference at the temporal encoding stage, we adopt a variable-aware strategy and encode each variable independently~\cite{PatchTST-2022,iTransformer-2023,RethinkingCI-2023,OracleAD-2025}.
Accordingly, each variable is processed by its own temporal encoder parameterized by $\theta_i$, allowing TENET to preserve variable-specific temporal characteristics before any cross-variable interaction is introduced.
For notational simplicity, within each window we re-index local time positions from $1$ to $L$, so that $\mathbf{x}^{(i)}_{1:L}$ denotes the univariate sequence of variable $i$ in $\mathbf{X}$.
For each variable $i$, the temporal layer $f_t(\cdot;\theta_i)$ maps the univariate window $\mathbf{x}^{(i)}_{1:L}$ to a current-point state token.
The key design choice is to encode temporal context according to its relevance to the target point, so that the temporal layer constructs the variable state at that point rather than summarizing the window uniformly over time.
In this sense, the temporal layer performs \emph{current-point-conditioned history retrieval}: the current point determines which historical information is most relevant for constructing the current-point variable state.
Operationally, $f_t$ is implemented as a stack of $M$ Temporal blocks, each of which updates the current-point anchor by retrieving history from the local window.

We first map $\mathbf{x}^{(i)}_{1:L}$ to value embeddings
$\mathcal{E}^{(i)}\in\mathbb{R}^{L\times d}$.
To encode temporal order relative to the current point, we use anchor-centered lag embeddings~\cite{RelativePE-2018}.
For each position $\ell\in\{1,\dots,L\}$, we define the lag $\tau_\ell = L-\ell$, so that $\tau_L=0$ corresponds to the current point and larger $\tau_\ell$ indicates farther history.
We then obtain temporal embeddings $\boldsymbol{\mathcal{T}}^{(i)}\in\mathbb{R}^{L\times d_t}$, where the lag embedding at position $\ell$ is denoted by $\boldsymbol{\mathcal{T}}^{(i)}_\ell=\mathbf{r}_{\tau_\ell}$ with $\mathbf{r}_{\tau_\ell}\in\mathbb{R}^{d_t}$.
The value and lag embeddings are concatenated to form the variable-specific tokens $\mathbf{U}^{(i)}=\big[\mathcal{E}^{(i)};\boldsymbol{\mathcal{T}}^{(i)}\big]\in\mathbb{R}^{L\times (d+d_t)}$.

The temporal layer is current-point-centered: it starts from the current point and retrieves recent history only to the extent that the history helps explain the present state.
Accordingly, we initialize the evolving anchor state by the embedding of the last time point, $\mathbf{z}^{(i)}_0 = \mathcal{E}^{(i)}_L \in \mathbb{R}^d$.
At Temporal block $m$, the core operation is an \emph{anchored cross-attention} module: the query is derived from the current anchor state, while keys and values are obtained from the full window after token-wise normalization:
\begin{equation}
\mathbf{q}^{(i)} = \mathrm{LN}_q\!\left(\mathbf{z}^{(i)}_{m-1}\right)\mathbf{W}_Q,\quad
\mathbf{K}^{(i)} = \mathrm{LN}_{kv}\!\left(\mathbf{U}^{(i)}\right)\mathbf{W}_{K},\quad
\mathbf{V}^{(i)} = \mathrm{LN}_{kv}\!\left(\mathbf{U}^{(i)}\right)\mathbf{W}_{V}.
\end{equation}
For head $r\in\{1,\dots,H_a\}$ with per-head dimension $d_h=d/H_a$, the attention weights and retrieved context are
\begin{equation}
\boldsymbol{\alpha}^{(i,r)}=\mathrm{softmax}\!\left(\frac{\mathbf{q}^{(i,r)}(\mathbf{K}^{(i,r)})^\top}{\sqrt{d_h}}\right)\in\mathbb{R}^{1\times L},
\qquad
\mathbf{h}^{(i,r)}=\boldsymbol{\alpha}^{(i,r)}\mathbf{V}^{(i,r)}\in\mathbb{R}^{1\times d_h},
\end{equation}
and the multi-head output is
$\mathbf{h}^{(i)}=
\begin{bmatrix}
\mathbf{h}^{(i,1)} & \cdots & \mathbf{h}^{(i,H_a)}
\end{bmatrix}
\mathbf{W}_O
\in \mathbb{R}^{d}$.
The retrieved context is merged with the anchor state through a \emph{gating mechanism}:
\begin{equation}
\begin{aligned}
\mathbf{g}^{(i)} &= \sigma\!\big(\mathbf{q}^{(i)}\mathbf{W}_{g} + \mathbf{b}_{g}\big), \\
\tilde{\mathbf{z}}^{(i)}_m &= \mathbf{g}^{(i)} \odot \mathbf{q}^{(i)} + (1-\mathbf{g}^{(i)}) \odot \mathbf{h}^{(i)} .
\end{aligned}
\label{eq:f_t}
\end{equation}
Here, $\sigma(u)=1/(1+e^{-u})$ is the element-wise sigmoid, and $\mathbf{g}^{(i)}\in(0,1)^d$ controls how much the model relies on the anchor state versus retrieved history.
The resulting representation is then refined by a Transformer-style feed-forward block with residual connection and layer normalization, $\mathbf{z}^{(i)}_m = \tilde{\mathbf{z}}^{(i)}_m + \mathrm{FFN}\!\left(\mathrm{LN}\!\left(\tilde{\mathbf{z}}^{(i)}_m\right)\right)$.
After $M$ layers, we apply a final layer normalization and obtain the current-point state token $\mathbf{e}^{(i)} = \mathrm{LN}\!\left(\mathbf{z}^{(i)}_M\right)$, which represents variable $i$ at the current point conditioned on its recent history.
This produces one current-point state token for each variable before inter-variable contextualization is introduced.
\vspace{-0.5em}
\paragraph{Inter-Variable Layer.}
Temporal modeling yields a set of current-point state tokens, $\mathbf{E} = [\mathbf{e}^{(1)}; \dots; \mathbf{e}^{(N)}]$.
We then model inter-variable dependencies by applying self-attention along the variable dimension.
With this ordering, TENET first constructs one current-point state token per variable and then models dependencies among variables.
Concretely, $f_v(\cdot)$ is implemented as a stack of Transformer-style blocks over the variable axis, each composed of multi-head self-attention followed by a position-wise feed-forward network with residual connections and layer normalization.
The final token-wise layer normalization is treated as part of this layer.
The output of the inter-variable module is the contextualized variable-level state set
\[
\mathbf{S} = f_v(\mathbf{E}) \in \mathbb{R}^{N \times d}.
\]
Each token is updated using information from the other variables, producing a structured variable-level state set $\mathbf{S}$.
This set is an intermediate representation.
It preserves one current-point state token for each variable after inter-variable contextualization, but it is not yet the final decision state.
\vspace{-0.5em}
\paragraph{Aggregation Layer.}
To obtain the current-point system state used for scoring, we pool the contextualized variable-level states in $\mathbf{S}$ into $\mathbf{c}\in\mathbb{R}^{h}$.
This aggregation design is inspired by the centralized core representation in recent multivariate time-series models~\cite{SOFTS-2024}.
We use summation so that variable-level contributions can accumulate in the system state, allowing weak deviations distributed across variables to remain visible after pooling:
\begin{equation}
\label{eq:f_g}
\mathbf{c}
=
f_g(\mathbf{S})
=
\rho\!\left(\sum_{i=1}^{N} \phi\!\left(\mathbf{s}^{(i)}\right)\right).
\end{equation}
Here, $\phi:\mathbb{R}^{d}\!\to\!\mathbb{R}^{h}$ is a two-layer MLP with GELU activation and $\rho:\mathbb{R}^{h}\!\to\!\mathbb{R}^{h}$ is a single linear layer followed by GELU.
We refer to $\mathbf{c}$ as the current-point system state.
This final decision state is used as the input to our simple one-class scoring objective.
\vspace{-0.5em}
\subsection{Training and Inference}

We train TENET only on normal windows and use the same compatibility score at both training and inference.
We deliberately adopt a simple fixed one-class objective because the practical goal of anomaly detection is not necessarily to recover the normal-state distribution as faithfully as possible, but to learn a state space that enables a useful compatibility ranking between normal and anomalous states.
From this perspective, fixing the reference distribution lets the experiment focus on state construction rather than density-model complexity.

With the fixed reference distribution
\(
p_{\mathrm{ref}}(\mathbf{c})=\mathcal{N}(\mathbf{0}, \mathbf{I}),
\)
we define the compatibility score as the negative log-likelihood (NLL) under this reference:
\begin{equation}
\label{eq:score_train_infer}
\mathcal{A}(\mathbf{X})
=
-\log p_{\mathrm{ref}}\!\left(f_{\Theta}(\mathbf{X})\right)
=
\frac{1}{2}\left\|f_{\Theta}(\mathbf{X})\right\|_2^2 + C,
\end{equation}
where $C$ is constant with respect to $\mathbf{X}$.
Training minimizes the average score over normal windows, encouraging TENET to place normal current-point states near the origin under the same score used at inference time.
At inference time, for each time step $t$ we form the window $\mathbf{X}_t=\mathbf{x}_{t-L+1:t}$ and assign the anomaly score of the current point $\mathbf{x}_t$ by $\mathcal{A}(\mathbf{X}_t)$.
The fixed NLL objective has a constant-map global optimum, so it is important to verify whether a detector preserves anomaly-ranking information without collapse over the finite training duration used in our experiments.
Additional discussion is provided in Appendix~\ref{appendix:state-geometry-rationale}, and an extended-training stability analysis is provided in Appendix~\ref{appendix:collapse_diagnosis}.
Appendix~\ref{appendix:gaussian-reference} compares fixed and learnable Gaussian references.

\section{Experiments}
\subsection{Experimental Setup}
\paragraph{Datasets.}
We evaluate our method on four multivariate time series anomaly detection benchmarks:
\textbf{PSM}~\cite{PSM-2021} (server monitoring, 25 variables),
\textbf{SMD}~\cite{OmniAnomaly-2019} (server machine, 38 variables),
\textbf{SWaT}~\cite{SWaT-2016} (water treatment, 51 variables),
and \textbf{SWAN}~\cite{SWAN-2020} (solar activity, 38 variables).
Dataset-selection rationale and statistics are provided in Appendix~\ref{appendix:datasets}.
\vspace{-0.3em}
\paragraph{Evaluation Metrics.}
Following the evaluation practices recommended in recent
studies~\cite{TSADEvaulation-2022, TimeSeAD-2023, TSAD_comprehensive_eval-2022, Elephant_in_the_Room-2024, TAB-2025}, we report performance using a comprehensive set of
evaluation metrics.
Specifically, we report point-wise ranking metrics AUC-ROC (A-ROC)~\cite{ROC-2006},
AUC-PR (A-PR)~\cite{PR_ROC-2006},
as well as range-aware, tolerance-aware metrics VUS-ROC (V-ROC) and VUS-PR (V-PR)~\cite{VUS-2022}.
Detailed definitions and evaluation protocols are provided in Appendix~\ref{appendix:metrics}.
\vspace{-0.3em}
\paragraph{Baselines and Encoders.}
The main benchmark comparison includes a broad set of MTSAD baselines spanning classical anomaly detection methods, latent reconstruction, representation learning, and forecasting-based approaches. Specifically, Table~\ref{tab:main} includes PCA~\cite{PCA-2003}, OCSVM~\cite{SVM_ND-1999}, Isolation Forest (IF)~\cite{Isolation_Forest-2008}, DAGMM~\cite{DAGMM-2018}, OmniAnomaly (Omni)~\cite{OmniAnomaly-2019}, Anomaly Transformer (A.T)~\cite{AnomalyTransformer-2021}, PatchTST~\cite{PatchTST-2022}, TimesNet~\cite{TimesNet-2022}, iTransformer (iTrans)~\cite{iTransformer-2023}, ModernTCN (Modern)~\cite{ModernTCN-2023}, SARAD~\cite{SARAD-2024}, CATCH~\cite{CATCH-2025}, DCdetector (DC)~\cite{DCdetector-2023}, CAROTS~\cite{CAROTS-2025}, THOC~\cite{THOC-2020}, TTM~\cite{TTMs-2024}, Timer~\cite{Timer-2024}, Chronos~\cite{Chronos-2024}, and CrossAD~\cite{CrossAD-2025}.
For the encoder--objective comparison in Table~\ref{tab:encoder_framework}, we compare TENET with five alternative sequence encoders: LSTM~\cite{EncDec-AD-2016}, Dilated RNN~\cite{THOC-2020}, Mamba~\cite{Mamba-2024}, ModernTCN~\cite{ModernTCN-2023}, and GPT4TS~\cite{GPT4TS-2023}. Each encoder produces a same-dimensional pooled state for the current point, allowing us to isolate the effect of the encoder under matched one-class objectives.
\vspace{-0.3em}
\paragraph{Implementation Details}
\label{sec:implementation}
For TENET, we construct model inputs as fixed-length sliding windows with length $L=20$.
The variable-token dimension is set to $d=256$, the pooled state dimension is set to $h=256$, and the temporal embedding dimension is set to $d_t=16$. 
The temporal layer $f_t$ is implemented with a depth of $M=2$ Temporal blocks and the inter-variable layer $f_v$ is implemented with depth 2.
All results in Table~\ref{tab:main} are our reruns using a TAB-based pipeline, not public leaderboard values. Methods share data loading, normalization, and evaluation, while baselines retain their recommended hyperparameters and training schedules. Differences from TAB's public protocol are detailed in Appendix~\ref{appendix:baselines-protocol}.
For TENET, we use a batch size of 1024 and train for 10 epochs on all four main benchmark datasets.
TENET is optimized with AdamW~\cite{AdamW-2017}.
For PSM, SWaT, and SWAN, we use learning rate $1\times10^{-6}$ and for SMD, we use learning rate $5\times10^{-5}$.

\subsection{Main Results}
\label{sec:main_results}
% ...

\begin{table*}[t]
\caption{Multivariate time series anomaly detection performance on real-world datasets. All metrics are reported in percent and averaged over five random seeds. \textbf{Bold} indicates the best score.}
\label{tab:main}
\centering
\setlength{\tabcolsep}{1.5pt}
\renewcommand{\arraystretch}{0.95}
\begin{scriptsize}
\begin{tabular}{c|cccc|cccc|cccc|cccc}
\toprule
\multicolumn{1}{c|}{\textbf{Dataset}} &
\multicolumn{4}{c|}{\textbf{PSM (25 vars)}} &
\multicolumn{4}{c|}{\textbf{SMD (38 vars)}} &
\multicolumn{4}{c|}{\textbf{SWaT (51 vars)}} &
\multicolumn{4}{c}{\textbf{SWAN (38 vars)}} \\
\midrule
\textbf{Metric} &
\textbf{A-ROC} & \textbf{A-PR} & \textbf{V-ROC} & \textbf{V-PR} &
\textbf{A-ROC} & \textbf{A-PR} & \textbf{V-ROC} & \textbf{V-PR} &
\textbf{A-ROC} & \textbf{A-PR} & \textbf{V-ROC} & \textbf{V-PR} &
\textbf{A-ROC} & \textbf{A-PR} & \textbf{V-ROC} & \textbf{V-PR} \\
\midrule

IF
& 54.23 & 33.41 & 56.47 & 37.78
& 56.59 & 15.63 & 56.21 & 15.91
& 36.35 & 10.27 & 41.37 & 10.17
& 48.96 & 38.12 & 52.67 & 62.27 \\
PCA
& 65.89 & 47.32 & 65.59 & 47.53
& 77.11 & 60.86 & 76.95 & 62.14
& 81.85 & 72.64 & 81.77 & 48.17
& 72.89 & 63.16 & 64.46 & 72.05 \\
OCSVM
& 69.29 & 50.83 & 69.12 & 50.97
& 76.87 & 61.71 & 76.52 & 61.42
& 81.76 & 71.80 & 68.54 & 42.38
& 71.36 & 60.75 & 63.99 & 70.90 \\
\midrule
DAGMM
& 54.91 & 32.55 & 54.80 & 32.05
& 78.25 & 59.90 & 77.73 & 19.91
& 61.96 & 16.47 & 36.61 & 9.67
& 67.18 & 55.59 & 60.08 & 64.16 \\
THOC
& 64.17 & 43.56 & 64.03 & 44.19
& 79.80 & 62.64 & 71.41 & 66.45
& 55.58 & 24.19 & 55.32 & 27.12
& 73.11 & 62.59 & 58.96 & 73.52 \\
\midrule
DC
& 49.86 & 27.78 & 50.11 & 27.89
& 50.31 & 10.15 & 49.97 & 10.05
& 49.89 & 12.14 & 50.00 & 11.92
& 49.74 & 32.54 & 50.00 & 55.05 \\
CAROTS
& 75.48 & 53.03 & 66.05 & 47.41
& 83.26 & 57.46 & 80.54 & 54.06
& 83.10 & 73.28 & 64.27 & 48.31
& 67.60 & 56.60 & \textbf{76.00} & 69.70 \\
\midrule

Omni
& 63.95 & 45.00 & 61.23 & 52.49
& \textbf{95.24} & 64.35 & 70.26 & 54.99
& 82.15 & 72.73 & 81.98 & 64.42
& 75.63 & 63.32 & 68.36 & 74.43 \\
A.T
& 38.35 & 24.28 & 50.96 & 49.76
& 48.44 & 10.73 & 50.24 & 33.60
& 42.58 & 11.93 & 50.72 & 17.00
& 41.69 & 31.27 & 50.14 & 63.13 \\
PatchTST
& 58.68 & 37.91 & 50.62 & 50.31
& 66.44 & 23.83 & 50.87 & 30.51
& 24.36 & 8.40  & 49.84 & 11.92
& 55.63 & 47.53 & 50.98 & 75.11 \\
TimesNet
& 59.09 & 38.95 & 50.67 & 49.86
& 65.18 & 23.37 & 51.22 & 31.05
& 28.97 & 10.71 & 20.80 & 51.20
& 70.85 & 60.71 & 50.48 & 75.41 \\
iTrans
& 59.22 & 38.27 & 51.23 & 44.54
& 67.03 & 24.49 & 50.91 & 28.38
& 24.57 & 8.33  & 49.90 & 12.05
& 52.10 & 42.11 & 53.27 & 69.31 \\
Modern
& 59.21 & 38.59 & 50.68 & 50.55
& 65.90 & 24.09 & 50.45 & 41.73
& 24.50 & 9.00  & 50.15 & 13.05
& 53.48 & 46.84 & 52.62 & 74.46 \\
SARAD
& 62.86 & 41.67 & 57.63 & 38.64
& 93.95 & 59.73 & 53.65 & 31.77
& 85.40 & 64.77 & 86.30 & 62.72
& 71.87 & 60.44 & 56.93 & 77.37 \\
CATCH
& 64.75 & 43.40 & 57.00 & 45.95
& 67.26 & 25.15 & 50.79 & 33.02
& 33.31 & 13.39 & 51.89 & 18.70
& 50.47 & 43.98 & 53.48 & 73.01 \\
CrossAD
& 61.79 & 41.30 & 50.61 & 50.67
& 87.53 & 65.74 & 50.92 & 46.09
& 55.92 & 14.28 & 50.32 & 13.88
& 61.88 & 52.51 & 53.64 & 74.78 \\
\midrule

TTM
& 53.99 & 33.83 & 51.29 & 45.30
& 59.64 & 21.41 & 50.30 & 52.28
& 26.68 & 11.55 & 51.20 & 15.58
& 64.76 & 52.77 & 53.04 & 74.34 \\
Timer
& 55.60 & 35.93 & 51.38 & 46.95
& 58.64 & 21.51 & 50.28 & 35.22
& 28.62 & 11.91 & 51.58 & 15.76
& 62.27 & 51.44 & 52.56 & 73.36 \\
Chronos
& 56.39 & 36.89 & 51.36 & 52.58
& 58.02 & 15.82 & 50.16 & 13.68
& 79.86 & 38.17 & 78.27 & 34.06
& 52.86 & 45.94 & 53.06 & 72.72 \\
GPT4TS
& 57.79 & 37.86 & 50.29 & 56.14
& 61.42 & 24.21 & 50.80 & 43.49
& 24.27 & 8.53 & 50.21 & 25.69
& 58.97 & 48.75 & 50.69 & 75.09 \\
\midrule

\textbf{TENET} & \textbf{86.14} & \textbf{74.27} & \textbf{85.82} & \textbf{74.21}
& 90.04 & \textbf{74.78} & \textbf{89.50} & \textbf{74.41}
& \textbf{87.78} & \textbf{79.63} & \textbf{87.67} & \textbf{79.45}
& \textbf{88.36} & \textbf{84.07} & 72.98 & \textbf{80.97} \\
\bottomrule
\end{tabular}
\end{scriptsize}
\end{table*}

\paragraph{Benchmark Results.}
Table~\ref{tab:main} summarizes the quantitative results on four benchmark datasets.
Across all four datasets, TENET with the fixed NLL-based compatibility score achieves strong performance on most of the reported metrics.
This result is notable because it shows that a time-point-centric state constructor paired with a simple fixed one-class objective can outperform detectors built around reconstruction or forecasting objectives.
In particular, TENET consistently surpasses strong reconstruction-based baselines such as CrossAD, forecasting-based baselines such as TTM, and latent one-class MTSAD baselines such as DAGMM and THOC.
On PSM, TENET improves V-PR by +21.72\%pt over OmniAnomaly and +42.16\%pt over DAGMM.
Compared with THOC, TENET improves V-PR by +30.02\%pt on PSM and +52.33\%pt on SWaT.
The gains on the range-aware VUS metrics are particularly notable in practice, because they indicate that TENET maintains elevated anomaly scores over anomalous segments rather than responding only at isolated timestamps.
Overall, the benchmark results indicate that one-class MTSAD can outperform observation-space discrepancy scoring when paired with time-point-centric representation modeling.


\begin{table*}[t]
\caption{Encoder and one-class objective comparison on PSM. All metrics are reported in percent and averaged over five random seeds. \textbf{Bold} indicates the best score.}
\label{tab:encoder_framework}
\centering
\setlength{\tabcolsep}{1.3pt}
\renewcommand{\arraystretch}{1.05}
\begin{scriptsize}
\begin{tabular}{c|cccc|cccc|cccc|cccc}
\toprule
\textbf{Objective}
& \multicolumn{4}{c|}{\textbf{NLL}} & \multicolumn{4}{c|}{\textbf{DeepSVDD}} & \multicolumn{4}{c|}{\textbf{THOC}} & \multicolumn{4}{c}{\textbf{DAGMM}} \\
\midrule
\textbf{Metric}
& \textbf{A-ROC} & \textbf{A-PR} & \textbf{V-ROC} & \textbf{V-PR}
& \textbf{A-ROC} & \textbf{A-PR} & \textbf{V-ROC} & \textbf{V-PR}
& \textbf{A-ROC} & \textbf{A-PR} & \textbf{V-ROC} & \textbf{V-PR}
& \textbf{A-ROC} & \textbf{A-PR} & \textbf{V-ROC} & \textbf{V-PR} \\
\midrule
LSTM
& 72.08 & 52.97 & 71.92 & 53.09
& 72.13 & 53.68 & 71.96 & 53.78
& 66.70 & 45.62 & 66.57 & 45.81
& 69.96 & 48.97 & 67.40 & 48.64 \\
Dilated RNN
& 66.02 & 46.83 & 65.89 & 47.02
& 72.90 & 53.30 & 72.70 & 53.39
& 60.10 & 40.33 & 60.04 & 40.57
& 65.83 & 44.88 & 65.59 & 45.06 \\
Mamba
& 64.98 & 40.87 & 64.89 & 41.10
& 70.61 & 46.75 & 70.47 & 46.92
& 66.51 & 40.53 & 66.41 & 40.80
& 71.15 & 48.26 & 70.87 & 48.36 \\
ModernTCN
& 74.39 & 51.69 & 63.85 & 49.53
& 74.23 & 51.55 & 63.92 & 49.45
& 78.86 & 54.22 & 78.80 & 55.00
& 73.75 & 47.78 & 60.59 & 44.59 \\
GPT4TS
& 74.67 & 45.58 & 61.04 & 42.00
& 66.47 & 40.99 & 60.57 & 41.44
& 73.27 & 43.21 & 60.03 & 39.69
& 44.91 & 31.85 & 44.94 & 32.34 \\
\midrule
\textbf{TENET}
& \textbf{86.14} & \textbf{74.27} & \textbf{85.82} & \textbf{74.21}
& \textbf{82.65} & \textbf{66.79} & \textbf{82.36} & \textbf{66.82}
& \textbf{80.95} & \textbf{65.36} & \textbf{80.67} & \textbf{65.41}
& \textbf{74.52} & \textbf{56.92} & \textbf{74.00} & \textbf{57.05} \\
\bottomrule
\end{tabular}
\end{scriptsize}
\vspace{-0.5em}
\end{table*}

\paragraph{One-Class Objective Generality.}
To examine whether TENET's advantage is specific to the default fixed NLL-based compatibility score, we construct a full encoder--objective matrix on PSM (Table~\ref{tab:encoder_framework}).
We compare six point-wise encoders under four representative one-class objectives, all applied to the same-dimensional pooled state $\mathbf{c} \in \mathbb{R}^{h}$ for the current point.
This setup separates the effects of encoder design from those of objective choice.

Table~\ref{tab:encoder_framework} shows that TENET ranks first under every objective and every metric.
Although the strongest non-TENET encoder varies across objectives, TENET remains consistently superior, with especially large margins under the fixed standard normal reference NLL and THOC.
This result shows that TENET's advantage is not explained merely by producing a current-point state.
Rather, under the same scoring setup, the key factor is how the current-point system state is constructed.
These results therefore support one of our core claims: in one-class MTSAD, constructing an appropriate current-point system state is a key factor behind strong detection performance.
Objective implementation details and additional encoder--objective results are provided in Appendices~\ref{appendix:oneclass_objectives} and~\ref{appendix:encoder_ext}, respectively.
\vspace{-0.5em}

\subsection{Representation Geometry Analysis}
\definecolor{tenetBlue}{RGB}{70,130,180}
\definecolor{lstmRed}{RGB}{220,0,30}
\definecolor{dilOrange}{RGB}{230,159,0}
\definecolor{mambaGreen}{RGB}{0,158,115}
\definecolor{tcnPurple}{RGB}{140,86,178}
\definecolor{gptRose}{RGB}{204,121,167}

\begin{figure*}[t]
\centering
\begin{tikzpicture}

\begin{groupplot}[
    group style={group size=4 by 1, horizontal sep=2.3em},
    width=0.302\linewidth,
    height=0.24\linewidth,
    ymajorgrids=true,
    grid style={dotted},
    x tick label style={font=\scriptsize},
    yticklabel style={font=\scriptsize},
    xlabel style={font=\scriptsize},
    ylabel style={font=\scriptsize},
    title style={font=\scriptsize, yshift=-6pt},
    every axis plot/.append style={thick, mark size=1.5pt},
]

\nextgroupplot[
    xmin=1, xmax=10,
    ymin=0, ymax=135,
    xtick={1,5,10},
    xlabel={Epoch},
    title={(a) eRank},
]
\addplot+[draw=tenetBlue, very thick, mark=none]
  table[col sep=comma, x=epoch, y=erank_tenet]{appendices/collapse_nll_encoder.csv};
\addplot+[draw=lstmRed, thick, opacity=0.7, mark=none]
  table[col sep=comma, x=epoch, y=erank_lstm]{appendices/collapse_nll_encoder.csv};
\addplot+[draw=dilOrange, thick, opacity=0.7, mark=none]
  table[col sep=comma, x=epoch, y=erank_dilatedrnn]{appendices/collapse_nll_encoder.csv};
\addplot+[draw=mambaGreen, thick, opacity=0.7, mark=none]
  table[col sep=comma, x=epoch, y=erank_mamba]{appendices/collapse_nll_encoder.csv};
\addplot+[draw=tcnPurple, thick, opacity=0.7, mark=none] coordinates {
    (1,59.3) (2,74.4) (3,87.0) (4,97.0) (5,103.7)
    (6,110.9) (7,115.7) (8,119.5) (9,122.4) (10,125.0)
};
\addplot+[draw=gptRose, solid, thick, opacity=0.7, mark=none] coordinates {
    (1,41.8) (2,63.2) (3,83.4) (4,105.5) (5,120.4) (6,96.6) (7,128.6) (8,121.4) (9,66.0) (10,45.7)
};

\nextgroupplot[
    xmin=1, xmax=100,
    ymin=0, ymax=1.05,
    xtick={1,20,40,60,80,100},
    xlabel={Components},
    title={(b) EVR},
]
\addplot+[draw=tenetBlue, very thick, mark=none]
  table[col sep=comma, x=k, y=evr_tenet]{appendices/evr_curves.csv};
\addplot+[draw=lstmRed, thick, opacity=0.7, mark=none]
  table[col sep=comma, x=k, y=evr_lstm]{appendices/evr_curves.csv};
\addplot+[draw=dilOrange, thick, opacity=0.7, mark=none]
  table[col sep=comma, x=k, y=evr_drnn]{appendices/evr_curves.csv};
\addplot+[draw=mambaGreen, thick, opacity=0.7, mark=none]
  table[col sep=comma, x=k, y=evr_mamba]{appendices/evr_curves.csv};
\addplot+[draw=tcnPurple, thick, opacity=0.7, mark=none]
  table[col sep=comma, x=k, y=evr_modern_tcn]{appendices/evr_curves.csv};
\addplot+[draw=gptRose, solid, thick, opacity=0.7, mark=none]
  table[col sep=comma, x=k, y=evr_gpt4ts]{appendices/evr_curves.csv};

\nextgroupplot[
    xmin=1, xmax=10,
    ymin=30, ymax=72,
    xtick={1,5,10},
    xlabel={Epoch},
    title={(c) Dir-Only V-PR (\%)},
]
\addplot+[draw=tenetBlue, very thick, mark=none] coordinates {
    (1,50.89) (2,57.71) (3,61.47) (4,63.93) (5,65.59)
    (6,66.80) (7,67.33) (8,67.62) (9,68.18) (10,68.61)
};
\addplot+[draw=lstmRed, thick, opacity=0.7, mark=none] coordinates {
    (1,47.06) (2,47.44) (3,47.69) (4,47.92) (5,48.12)
    (6,48.24) (7,48.30) (8,48.36) (9,48.31) (10,48.19)
};
\addplot+[draw=dilOrange, thick, opacity=0.7, mark=none] coordinates {
    (1,43.29) (2,43.45) (3,43.68) (4,43.71) (5,43.79)
    (6,43.85) (7,43.88) (8,44.00) (9,43.96) (10,43.95)
};
\addplot+[draw=mambaGreen, thick, opacity=0.7, mark=none] coordinates {
    (1,44.17) (2,44.85) (3,44.73) (4,45.32) (5,45.22)
    (6,46.04) (7,46.13) (8,46.50) (9,46.69) (10,48.55)
};
\addplot+[draw=tcnPurple, thick, opacity=0.7, mark=none] coordinates {
    (1,34.41) (2,35.09) (3,35.98) (4,36.66) (5,37.55)
    (6,37.43) (7,38.98) (8,39.68) (9,39.49) (10,38.45)
};
\addplot+[draw=gptRose, solid, thick, opacity=0.7, mark=none] coordinates {
    (1,37.28) (2,36.28) (3,41.14) (4,49.10) (5,47.16) (6,44.68) (7,46.89) (8,43.90) (9,44.16) (10,44.97)
};

\nextgroupplot[
    xmin=1, xmax=10,
    ymin=30, ymax=80,
    xtick={1,5,10},
    xlabel={Epoch},
    title={(d) Norm-Only V-PR (\%)},
]
\addplot+[draw=tenetBlue, very thick, mark=none] coordinates {
    (1,55.43) (2,60.59) (3,63.69) (4,66.04) (5,68.19)
    (6,70.01) (7,71.51) (8,72.52) (9,73.08) (10,73.37)
};
\addplot+[draw=lstmRed, thick, opacity=0.7, mark=none] coordinates {
    (1,47.56) (2,47.81) (3,48.06) (4,48.28) (5,48.56)
    (6,48.86) (7,49.20) (8,49.59) (9,50.01) (10,50.46)
};
\addplot+[draw=dilOrange, thick, opacity=0.7, mark=none] coordinates {
    (1,48.60) (2,49.70) (3,50.54) (4,51.08) (5,51.32)
    (6,51.42) (7,51.25) (8,50.86) (9,50.29) (10,49.47)
};
\addplot+[draw=mambaGreen, thick, opacity=0.7, mark=none] coordinates {
    (1,34.87) (2,36.37) (3,37.89) (4,39.35) (5,40.64)
    (6,41.72) (7,42.61) (8,43.37) (9,43.98) (10,44.50)
};
\addplot+[draw=tcnPurple, thick, opacity=0.7, mark=none] coordinates {
    (1,48.85) (2,48.46) (3,48.75) (4,48.81) (5,49.08)
    (6,49.40) (7,49.51) (8,49.42) (9,49.77) (10,50.33)
};
\addplot+[draw=gptRose, solid, thick, opacity=0.7, mark=none] coordinates {
    (1,45.23) (2,45.84) (3,46.93) (4,46.65) (5,46.88) (6,47.65) (7,46.82) (8,47.03) (9,45.59) (10,45.19)
};

\end{groupplot}

\node[anchor=north, yshift=0.2em] at (current bounding box.south) {%
\scriptsize
\begin{tabular}{@{}c@{\hspace{0.55em}}c@{\hspace{0.55em}}c@{\hspace{0.55em}}c@{\hspace{0.55em}}c@{\hspace{0.55em}}c@{}}
\tikz{\draw[tenetBlue, very thick] (0,0) -- (0.4,0);}~TENET &
\tikz{\draw[lstmRed, thick, opacity=0.7] (0,0) -- (0.4,0);}~LSTM &
\tikz{\draw[dilOrange, thick, opacity=0.7] (0,0) -- (0.4,0);}~Dilated RNN &
\tikz{\draw[mambaGreen, thick, opacity=0.7] (0,0) -- (0.4,0);}~Mamba &
\tikz{\draw[tcnPurple, thick, opacity=0.7] (0,0) -- (0.4,0);}~ModernTCN &
\tikz{\draw[gptRose, solid, thick, opacity=0.7] (0,0) -- (0.4,0);}~GPT4TS
\end{tabular}%
};
\end{tikzpicture}

\caption{
Representation diagnostics on PSM across TENET and alternative encoders.
(a) Effective rank across epochs, where higher values indicate a more distributed representation.
(b) Cumulative explained variance ratio at the final epoch, where a slower rise indicates weaker spectral concentration.
(c) V-PR using only the directional component of the state vector, where higher values indicate stronger anomaly-relevant angular structure.
(d) V-PR using only the radial component of the state vector, where higher values indicate stronger anomaly-relevant radial structure.
}
\label{fig:collapse_combined}
\vspace{-1em}
\end{figure*}
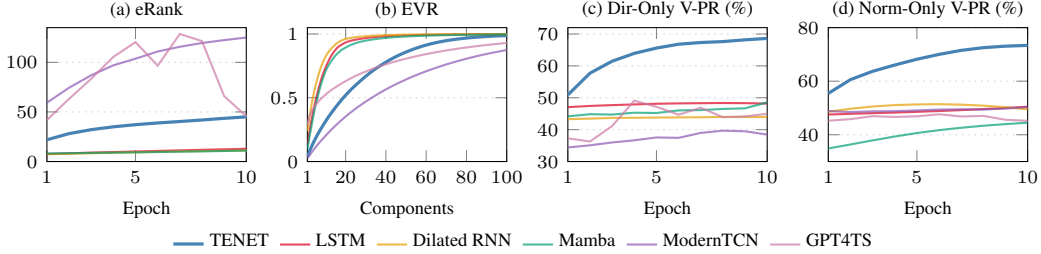

A useful one-class state space must resolve a central tension. One-class learning puts pressure on normal states to contract toward a reference such as the origin or a hypersphere center, but when this contraction becomes too strong, the representation can collapse and lose information needed to distinguish anomalies. Figure~\ref{fig:collapse_combined} examines this tension under the same NLL objective on PSM. We use four complementary diagnostics. Effective rank~\cite{EffectiveRank-2007} measures how many latent directions are substantially used, while cumulative explained variance ratio (EVR) measures how quickly variance concentrates in the leading principal directions. Direction-only and norm-only probes recompute V-PR after scoring only angular deviation or radial magnitude, respectively, to test which parts of the state retain anomaly-discriminative information.

The first two panels show that TENET avoids the low-rank collapse observed in several alternative encoders. In Figure~\ref{fig:collapse_combined}(a), TENET stabilizes near an effective rank of ${\sim}69$, whereas several alternatives contract to much lower-rank representations at the final epoch. Figure~\ref{fig:collapse_combined}(b) shows the same pattern spectrally: TENET spreads variance much more gradually, reaching only 31\% cumulative EVR in the top 10 components, while lower-rank alternatives already concentrate most variance in a small number of directions. However, broad geometry alone is not the target. ModernTCN and GPT4TS also occupy relatively broad subspaces, yet their anomaly detection performance remains weaker than TENET. The lesson from Figure~\ref{fig:collapse_combined} is therefore not that a wider representation is automatically better, but that the representation must be distributed in a way that remains aligned with anomaly scoring.
The last two panels clarify what this alignment means. In Figure~\ref{fig:collapse_combined}(c), TENET reaches 68.6\% V-PR using only directional information, indicating that anomaly-relevant angular structure is retained in the learned state. In Figure~\ref{fig:collapse_combined}(d), TENET also remains strongest under a norm-only score, showing that radial information is organized as well. Although the deployed fixed Gaussian NLL is radial, these probes suggest that TENET does not merely learn a trivial norm separation. Instead, its encoder organizes both radial and angular structure so that anomaly-relevant variation remains separable in the learned state space.
Detailed definitions of diagnostics are provided in Appendix~\ref{appendix:training_stability}.
\vspace{-0.5em}

\subsection{Temporal Layer Variation}
\begin{table*}[t]
\caption{Architectural variation of the temporal layer within TENET on four benchmark datasets. All metrics are reported in percent and averaged over five random seeds. \textbf{Bold} indicates the best score.}
\label{tab:variation}
\centering
\setlength{\tabcolsep}{1pt}
\renewcommand{\arraystretch}{1.05}
\begin{scriptsize}
\begin{tabular}{c|cccc|cccc|cccc|cccc}
\toprule
\multicolumn{1}{c|}{\textbf{Dataset}} &
\multicolumn{4}{c|}{\textbf{PSM (25 vars)}} &
\multicolumn{4}{c|}{\textbf{SMD (38 vars)}} &
\multicolumn{4}{c|}{\textbf{SWaT (51 vars)}} &
\multicolumn{4}{c}{\textbf{SWAN (38 vars)}} \\
\midrule
\textbf{Metric} &
\textbf{A-ROC} & \textbf{A-PR} & \textbf{V-ROC} & \textbf{V-PR} &
\textbf{A-ROC} & \textbf{A-PR} & \textbf{V-ROC} & \textbf{V-PR} &
\textbf{A-ROC} & \textbf{A-PR} & \textbf{V-ROC} & \textbf{V-PR} &
\textbf{A-ROC} & \textbf{A-PR} & \textbf{V-ROC} & \textbf{V-PR} \\
\midrule
LSTM
& 66.58 & 49.38 & 66.47 & 49.57
& 84.70 & 69.75 & 84.07 & 69.84
& 82.40 & 72.47 & 82.31 & 68.80
& 63.54 & 54.41 & 59.46 & 67.99 \\
Dilated RNN
& 66.64 & 49.36 & 66.52 & 49.53
& 84.91 & 69.77 & 84.16 & 69.77
& 82.40 & 72.59 & 82.30 & 68.76
& 64.19 & 55.41 & 59.47 & 68.10 \\
Mamba
& 76.42 & 60.64 & 76.24 & 60.72
& 88.97 & 68.62 & 88.42 & 68.35
& 82.74 & 62.21 & 82.66 & 61.69
& 80.08 & 68.76 & 68.77 & 74.58 \\
\midrule
\textbf{TENET}
& \textbf{86.14} & \textbf{74.27} & \textbf{85.82} & \textbf{74.21}
& \textbf{90.04} & \textbf{74.78} & \textbf{89.50} & \textbf{74.41}
& \textbf{87.78} & \textbf{79.63} & \textbf{87.67} & \textbf{79.45}
& \textbf{88.36} & \textbf{84.07} & \textbf{72.98} & \textbf{80.97} \\
\bottomrule
\end{tabular}
\end{scriptsize}
\vspace{-0.5em}
\end{table*}

Table~\ref{tab:variation} reports results when the temporal layer in TENET is replaced with alternative sequence encoders while the rest of the model is kept fixed.
Specifically, we replace $f_t$ with LSTM, Dilated RNN, and Mamba, while keeping $f_v$, $f_g$, and the scoring head unchanged.
Under this controlled substitution, TENET consistently outperforms all alternatives across the four datasets.
Among the replacement architectures, Mamba is generally the strongest baseline, whereas LSTM and Dilated RNN show the largest overall degradation.
These trends suggest that generic recurrent and state-space encoders can capture useful local dynamics, but still fall short of TENET's current-point-conditioned temporal retrieval.
Overall, these results show that TENET's temporal layer is not an interchangeable backbone module.
For point-wise anomaly detection, temporal context should be retrieved with respect to the current point, not passively compressed into a generic sequence summary.

\vspace{-0.5em}

\subsection{Component Ablation}

\begin{table*}[t]
\caption{Component ablation study of TENET on four benchmark datasets. All metrics are reported in percent and averaged over five random seeds. \textbf{Bold} indicates the best score.}
\label{tab:model_analysis}
\centering
\setlength{\tabcolsep}{0.8pt}
\renewcommand{\arraystretch}{1.05}
\begin{tiny}
\begin{tabular}{cccc|cccc|cccc|cccc|cccc}
\toprule
\multicolumn{4}{c|}{\textbf{Components}} &
\multicolumn{4}{c|}{\textbf{PSM (25 vars)}} &
\multicolumn{4}{c|}{\textbf{SMD (38 vars)}} &
\multicolumn{4}{c|}{\textbf{SWaT (51 vars)}} &
\multicolumn{4}{c}{\textbf{SWAN (38 vars)}} \\
\midrule
\makecell{\textbf{Gating}\\\textbf{Mech.}}
& \makecell{\textbf{Lag}\\\textbf{Embed.}}
& \makecell{\textbf{Per-Var.}\\\textbf{Encoder}}
& \makecell{\textbf{Cross-Var.}\\\textbf{Attn.}}
& \textbf{A-ROC} & \textbf{A-PR} & \textbf{V-ROC} & \textbf{V-PR}
& \textbf{A-ROC} & \textbf{A-PR} & \textbf{V-ROC} & \textbf{V-PR}
& \textbf{A-ROC} & \textbf{A-PR} & \textbf{V-ROC} & \textbf{V-PR}
& \textbf{A-ROC} & \textbf{A-PR} & \textbf{V-ROC} & \textbf{V-PR} \\
\midrule
\ding{55} & \ding{51} & \ding{51} & \ding{51}
& 82.07 & 66.90 & 81.81 & 66.91
& 90.27 & 73.66 & 89.71 & 73.29
& 86.51 & 77.07 & 86.42 & 76.60
& 78.72 & 77.44 & 64.44 & 76.06 \\
\ding{51} & \ding{55} & \ding{51} & \ding{51}
& 82.30 & 67.55 & 82.01 & 67.56
& 88.83 & 72.90 & 88.31 & 72.60
& 86.08 & 74.71 & 85.98 & 74.39
& 87.22 & 81.97 & 72.10 & 79.97 \\
\ding{51} & \ding{51} & \ding{55} & \ding{51}
& 54.29 & 32.18 & 54.05 & 32.51
& 70.49 & 30.22 & 70.29 & 30.49
& 81.88 & 72.34 & 81.84 & 70.17
& 83.33 & 77.01 & 68.97 & 76.91 \\
\ding{51} & \ding{51} & \ding{51} & \ding{55}
& 84.70 & 72.47 & 84.36 & 72.41
& 89.76 & 74.49 & 89.22 & 74.03
& 87.60 & 79.09 & 87.49 & 78.75
& 88.18 & 83.93 & 72.91 & 80.96 \\
\midrule
\ding{51} & \ding{51} & \ding{51} & \ding{51}
& \textbf{86.14} & \textbf{74.27} & \textbf{85.82} & \textbf{74.21}
& \textbf{90.04} & \textbf{74.78} & \textbf{89.50} & \textbf{74.41}
& \textbf{87.78} & \textbf{79.63} & \textbf{87.67} & \textbf{79.45}
& \textbf{88.36} & \textbf{84.07} & \textbf{72.98} & \textbf{80.97} \\
\bottomrule
\end{tabular}
\end{tiny}
\vspace{-1em}
\end{table*}

Table~\ref{tab:model_analysis} summarizes ablations of the core components in TENET.
Removing the gating mechanism in $f_t$ degrades overall performance across the datasets, especially on the range-aware metrics.
In this variant, the variable token is formed solely from the attention-retrieved temporal context, without an explicit input-dependent gate between the anchor state and retrieved history.
The resulting drop (e.g., $-7.3$\%pt V-PR on PSM and $-2.9$\%pt on SWaT) indicates that temporal evidence should not be incorporated uniformly, but should instead be modulated in a variable- and input-dependent manner.

Replacing the anchor-centered lag embedding with standard sinusoidal positional encoding also reduces performance, suggesting that temporal order is more effective when encoded relative to the current point rather than by absolute position in the window.
The largest degradation occurs when independent per-variable encoders are replaced by a single shared encoder.
On PSM, this change lowers V-PR by 41.7\%pt, which is by far the largest loss among all ablations.
This supports the importance of variable-wise temporal modeling in TENET.
Cross-variable attention has a more modest effect, but it provides additional refinement by composing the variable-level states.
These ablation results support our design choice of TENET.

\vspace{-0.5em}
\subsection{Computational Scalability}
\label{sec:computational_scalability}
\begin{table}[t]
\caption{Computational scalability of TENET on synthetic data with $d=256$. Latency is reported in milliseconds(ms) and memory in gigabytes(GB).}
\label{tab:scalability}
\vspace{3pt}
\centering
\setlength{\tabcolsep}{2.0pt}
\renewcommand{\arraystretch}{1.05}
\begin{scriptsize}
\begin{tabular}{c|cc|cc|cc|cc|cc|cc|cc|cc}
\toprule
\textbf{Window}
& \multicolumn{4}{c|}{\textbf{L=20}} & \multicolumn{4}{c|}{\textbf{L=50}} & \multicolumn{4}{c|}{\textbf{L=100}} & \multicolumn{4}{c}{\textbf{L=150}} \\
\midrule
\textbf{Phase}
& \multicolumn{2}{c|}{\textbf{Inference}} & \multicolumn{2}{c|}{\textbf{Training}}
& \multicolumn{2}{c|}{\textbf{Inference}} & \multicolumn{2}{c|}{\textbf{Training}}
& \multicolumn{2}{c|}{\textbf{Inference}} & \multicolumn{2}{c|}{\textbf{Training}}
& \multicolumn{2}{c|}{\textbf{Inference}} & \multicolumn{2}{c}{\textbf{Training}} \\
\midrule
\textbf{Metric}
    & \textbf{Lat.} & \textbf{Mem.} & \textbf{Lat.} & \textbf{Mem.}
    & \textbf{Lat.} & \textbf{Mem.} & \textbf{Lat.} & \textbf{Mem.}
    & \textbf{Lat.} & \textbf{Mem.} & \textbf{Lat.} & \textbf{Mem.}
    & \textbf{Lat.} & \textbf{Mem.} & \textbf{Lat.} & \textbf{Mem.} \\
\midrule
N=25  & 8.30  & 0.20 & 39.30  & 1.01  & 8.30  & 0.57 & 39.90  & 1.48  & 8.40  & 0.58 & 40.10  & 2.25  & 8.40  & 0.58 & 50.00   & 3.02 \\
N=100 & 32.70 & 1.48 & 186.70 & 3.97  & 32.80 & 2.18 & 189.70 & 5.82  & 32.90 & 2.18 & 188.70 & 8.93  & 32.80 & 2.19 & 216.40  & 11.97 \\
N=400 & 143.20& 5.82 & 883.90 & 16.73 & 144.40& 8.60 & 906.30 & 24.14 & 146.70& 8.62 & 910.10 & 36.55 & 148.40& 8.64 & 1020.20 & 48.75 \\
\bottomrule
\end{tabular}
\end{scriptsize}
\vspace{-0.5em}
\end{table}

For fixed hidden dimension, TENET has two main per-window computational terms: $\mathcal{O}(NL)$ from the temporal layer and $\mathcal{O}(N^2)$ from the inter-variable layer.
Table~\ref{tab:scalability} reports measured latency and peak memory for inference (batch size 1) and training (batch size 64) on synthetic data over an $N \times L$ grid.
The measurements show that inference latency is primarily sensitive to the number of variables and changes only weakly with the window length in the measured range.
Across the tested grid, the observed growth with $N$ is closer to linear than quadratic, indicating that the inter-variable attention term is not the sole runtime bottleneck at these problem sizes.
Training memory includes parameters, gradients, optimizer states, and activations retained for backpropagation, and grows with both $N$ and $L$.
Training latency is mainly sensitive to $N$ with moderate additional dependence on $L$.
To examine memory use when both $N$ and $L$ are large, we profile $N=L=200$ with $d=256$, batch size 1, and FP32 precision.
Parameter storage is 1,394.36\,MB, inference adds 5.27\,MB over the pre-forward memory baseline, and peak memory is 1,409.82\,MB.
Thus, parameters account for most inference memory in this setting.
System configuration details are provided in Appendix~\ref{appendix:system-configuration}.
A cross-model cost comparison is provided in Appendix~\ref{appendix:cross-model-cost}.

\vspace{-0.5em}
\subsection{Hyperparameter Sensitivity}
\begin{figure*}[t]
\centering
\begin{tikzpicture}
\begin{groupplot}[
    group style={group size=4 by 1, horizontal sep=2em},
    width=0.29\linewidth,
    height=0.22\linewidth,
    ymajorgrids=true,
    grid style={dotted},
    enlarge x limits=0.1,
    x tick label style={font=\scriptsize},
    yticklabel style={font=\scriptsize},
    xlabel style={font=\scriptsize},
    ylabel style={font=\scriptsize},
    title style={font=\scriptsize, yshift=-6pt},
    every axis plot/.append style={thick, mark size=1.5pt},
]

\nextgroupplot[
    symbolic x coords={20,40,60,80,100,400},
    xtick={20,40,60,80,100,400},
    xticklabels={20,40,60,80,100,400},
    ymin=76, ymax=84,
    ylabel={V-PR (\%)},
    title={(a) Window Size $L$},
]
\addplot[name path=upper, draw=none, forget plot]
  coordinates {(20,81.69) (40,80.33) (60,81.35) (80,81.38) (100,81.94) (400,82.36)};
\addplot[name path=lower, draw=none, forget plot]
  coordinates {(20,80.25) (40,79.99) (60,79.15) (80,79.76) (100,79.86) (400,80.58)};
\addplot[fill=blue!15, forget plot] fill between[of=upper and lower];
\addplot+[draw=blue!70!black, mark=*, mark options={fill=blue!70!black}]
  coordinates {(20,80.97) (40,80.16) (60,80.25) (80,80.57) (100,80.90) (400,81.47)};

\nextgroupplot[
    symbolic x coords={64,128,256,512,1024},
    xtick={64,128,256,512,1024},
    xticklabels={64,128,256,512,1K},
    ymin=76, ymax=84,
    ylabel={},
    title={(b) Batch Size $B$},
]
\addplot[name path=upper, draw=none, forget plot]
  coordinates {(64,82.41) (128,81.75) (256,81.42) (512,81.55) (1024,81.69)};
\addplot[name path=lower, draw=none, forget plot]
  coordinates {(64,81.49) (128,81.03) (256,80.30) (512,80.15) (1024,80.25)};
\addplot[fill=blue!15, forget plot] fill between[of=upper and lower];
\addplot+[draw=blue!70!black, mark=*, mark options={fill=blue!70!black}]
  coordinates {(64,81.95) (128,81.39) (256,80.86) (512,80.85) (1024,80.97)};

\nextgroupplot[
    symbolic x coords={32,64,128,256,512},
    xtick={32,64,128,256,512},
    xticklabels={32,64,128,256,512},
    ymin=70, ymax=84,
    ylabel={},
    title={(c) Embedding Dim $d$},
]
\addplot[name path=upper, draw=none, forget plot]
  coordinates {(32,77.59) (64,80.03) (128,80.60) (256,81.69) (512,81.81)};
\addplot[name path=lower, draw=none, forget plot]
  coordinates {(32,70.21) (64,76.61) (128,78.68) (256,80.25) (512,80.73)};
\addplot[fill=blue!15, forget plot] fill between[of=upper and lower];
\addplot+[draw=blue!70!black, mark=*, mark options={fill=blue!70!black}]
  coordinates {(32,73.90) (64,78.32) (128,79.64) (256,80.97) (512,81.27)};

\nextgroupplot[
    symbolic x coords={16,32,64,128,256},
    xtick={16,32,64,128,256},
    xticklabels={16,32,64,128,256},
    ymin=70, ymax=84,
    ylabel={},
    title={(d) Pooling Dim $h$},
]
\addplot[name path=upper, draw=none, forget plot]
  coordinates {(16,79.94) (32,81.08) (64,81.04) (128,81.74) (256,81.69)};
\addplot[name path=lower, draw=none, forget plot]
  coordinates {(16,77.70) (32,77.86) (64,76.92) (128,78.76) (256,80.25)};
\addplot[fill=blue!15, forget plot] fill between[of=upper and lower];
\addplot+[draw=blue!70!black, mark=*, mark options={fill=blue!70!black}]
  coordinates {(16,78.82) (32,79.47) (64,78.98) (128,80.25) (256,80.97)};

\end{groupplot}
\end{tikzpicture}
\caption{
Hyperparameter sensitivity on SWAN.
Lines show V-PR (\%) averaged over five seeds and shaded regions indicate ${\pm}1$ standard deviation.
X-axis settings are equally spaced.
}
\label{fig:sensitivity}
\vspace{-1em}
\end{figure*}
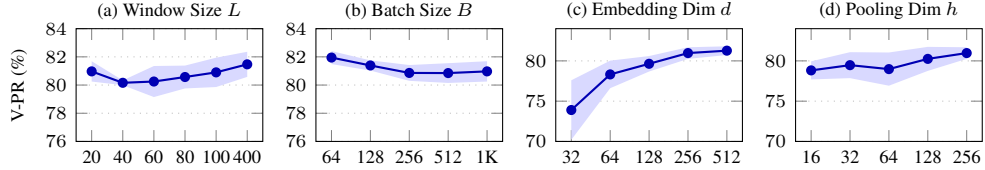

Figure~\ref{fig:sensitivity} illustrates the sensitivity of TENET to four hyperparameters on SWAN.
Overall, performance remains stable across a broad range of settings, indicating that TENET is not highly sensitive to moderate hyperparameter changes.
Window-size results remain stable from $L=20$ to $L=400$, while batch-size changes have little influence on performance.
For $L=400$, we use micro-batches of 128 with eight-step gradient accumulation for an effective batch size of 1024.
The embedding dimension $d$ and pooling dimension $h$ show clearer effects than window size or batch size, but in both cases performance stabilizes once a moderate capacity is reached.
This suggests that TENET is robust to hyperparameter choice and does not require delicate tuning to achieve strong performance.
Additional sensitivity tests under training contamination and test-time current-point perturbations are provided in Appendix~\ref{appendix:contam}.
\vspace{-0.5em}

\section{Conclusion}
\label{sec:conclusion}
We presented TENET, a current-point-centric encoder for multivariate time series anomaly detection.
TENET is motivated by the view that one-class MTSAD should not only ask how normality is scored, but also what representation is being scored.
Since MTSAD decisions are made for specific time points, TENET treats the input window not as the object to reconstruct, forecast, summarize, or score, but as decision-relevant context for constructing the current-point system state.

Across benchmark datasets, TENET achieves strong detection performance with a simple fixed NLL-based compatibility score.
Controlled comparisons further suggest that its advantage comes from decision-aligned representation construction rather than from a particular scoring rule.
Our findings suggest that one-class MTSAD benefits from interpreting the decision target as a time point and designing time-point encoders that construct the state to be scored.
Additional discussion of limitations and future directions is provided in the appendix~\ref{appendix:limitations}.

\bibliographystyle{plainnat}
\bibliography{reference}

\newpage
\appendix
\input{appendices/appendix_title}
\section{Design Rationale for TENET}
\label{appendix:state-geometry-rationale}

This appendix provides an informal design-level interpretation of why TENET's current-point-centric state constructor can produce the distributed and anomaly-discriminative geometry observed in Figure~\ref{fig:collapse_combined}.
The fixed standard-normal NLL used in the main experiments is intentionally simple:
\[
A(\mathbf{X})
=
\frac{1}{2}\|f_\Theta(\mathbf{X})\|_2^2+C .
\]
We use this score as a fixed compatibility interface rather than as a new one-class objective or a calibrated density estimator.
Its role is to keep the scoring geometry simple so that the empirical effect of the state constructor can be isolated.
As with compact one-class objectives based on a fixed reference, the loss can admit degenerate constant solutions in principle.

The operational goal in unsupervised MTSAD is not necessarily to recover the full normal-state density, but to learn an anomaly score whose test-time ordering ranks abnormal current-point states above normal ones.
Under the fixed NLL score, this ordering is induced by the radial magnitude of the learned current-point system state.
The relevant empirical question is therefore whether normal-only training with this simple compatibility score produces nontrivial states whose inference-time scores separate normal and anomalous points.
The benchmark results in Section~\ref{sec:main_results} evaluate this directly with threshold-free ranking metrics, and the encoder--objective comparisons in Table~\ref{tab:encoder_framework} and Appendix~\ref{appendix:encoder_ext} test whether TENET's advantage persists beyond the default NLL score.

TENET is designed around the hypothesis that this ranking depends primarily on what state is constructed before scoring.
Its temporal layer first forms one current-point token per variable rather than compressing the whole multivariate window into a single generic summary.
This variable-wise temporal construction reduces early cross-variable interference and allows different variables to retain their own temporal patterns before multivariate composition.
The shared-encoder ablation in Table~\ref{tab:model_analysis} supports this interpretation, although it also changes model capacity and should therefore be viewed as evidence for, not a complete isolation of, variable-specific specialization.

The temporal block retrieves history with respect to the current point.
At each block, the query is derived from the evolving current-point anchor, while keys and values come from the window context.
Thus, the model can select historical information according to its relevance to the present state rather than applying a fixed window summarization rule.
Anchor-centered lag embeddings express time relative to the current decision point, and the gate adaptively mixes the current anchor path with the retrieved-history path.
These components are intended to preserve information that helps explain the current point while avoiding uniform accumulation of all past observations.

After variable-wise temporal construction, TENET applies cross-variable attention over the resulting current-point tokens.
This ordering is important because inter-variable modeling is performed after each variable has already selected its relevant temporal context.
The aggregation layer then maps the structured variable-level state set into a single current-point system state for scoring.
Consequently, the fixed NLL is not applied to a raw window summary, but to a state produced by variable-wise temporal retrieval, cross-variable contextualization, and nonlinear set aggregation.

This design rationale is consistent with the representation diagnostics in Figure~\ref{fig:collapse_combined}.
TENET learns a state space that is less spectrally concentrated than several alternative encoders and retains anomaly-discriminative information under both radial and directional probes.
The deployed NLL score uses only radial magnitude, so the directional probe is diagnostic rather than part of the detector.
Its purpose is to show that the learned representation is not merely a trivial scalar norm separation.
Overall, TENET's advantage should be understood as coming from decision-aligned current-point state construction rather than from making the one-class objective more complex.

\newpage
\section{Extended-Training Stability}
\label{appendix:collapse_diagnosis}

The main TENET results use the fixed NLL objective in Eq.~\eqref{eq:score_train_infer}, which encourages compact current-point states.
We therefore include an extended-training stability analysis to assess the learned representation under prolonged optimization.
Specifically, we train TENET on PSM for 100 epochs, which is ten times longer than the default 10-epoch protocol used in the main experiments, and log per-epoch diagnostics.
Figure~\ref{fig:diagnosis_100ep} reports eight quantities side by side: (a) test mean state norm, (b) RMS pairwise distance among pooled states, (c) effective rank of the pooled-state covariance, (d) the cumulative explained-variance ratio at the final epoch, and (e)--(h) the four detection metrics (A-ROC, A-PR, V-ROC, V-PR) on the test split.
Test mean state norm and RMS pairwise distance diagnose compactness, where decreasing values reflect a more compact pooled-state distribution.
Effective rank~\cite{EffectiveRank-2007} and cumulative EVR diagnose whether increasing compactness is accompanied by low-rank behavior, either by reducing the number of active covariance directions or by concentrating variance in a few principal components.
The detection metrics track whether the changing representation still preserves anomaly-ranking information.

\paragraph{Compactness and Detection Performance.}
As expected from a one-class compactness objective, the pooled states become increasingly compact around the origin.
This is visible in the continued decrease of the mean state norm and the RMS pairwise distance among pooled states.
Figure~\ref{fig:diagnosis_100ep}(a) further shows that normal and anomalous states move toward smaller norms together, but their norm gap remains visible across training.
Thus, increasing compactness does not immediately erase the anomaly-ranking signal used by the fixed NLL score.
All four detection metrics continue to improve after the default epoch-10 checkpoint, reach their highest values later in training, and remain strong even under much longer training.
This suggests that TENET retains useful anomaly-ranking information under the fixed NLL objective, even as the pooled states become more compact.

\paragraph{Compactness and Representation Diversity.}
The representation diagnostics do not indicate simple low-rank collapse under extended training.
Effective rank continues to grow, and the cumulative EVR curve shows that the remaining variance is distributed across many directions rather than concentrated in only a few components.
Thus, prolonged NLL training mainly induces compactness-driven scale reduction and latent-state drift, while the learned state remains informative for anomaly ranking.

\input{figure/fig_diagnosis_100ep}

\newpage
\section{Pseudocode}
\label{appendix:tenet_pseudocode}

This appendix presents implementation-level pseudocode for TENET. The notation follows the PyTorch execution path used in our experiments. A dataloader returns a minibatch $\mathbf{X}_{\mathrm{raw}}\in\mathbb{R}^{B\times L\times N}$, where $B$ is the batch size, $L$ is the window length, and $N$ is the number of variables. The batch is transposed to $\mathbf{X}\in\mathbb{R}^{B\times N\times L}$ so that each variable can be encoded independently.

\begin{algorithm}[H]
\caption{TENET pseudocode}
\label{alg:tenet_forward}
\small
\begin{algorithmic}[1]
\STATE \textbf{Input:} Minibatch $\mathbf{X}\in\mathbb{R}^{B\times N\times L}$, variable-specific temporal encoders $\{f_t^{(i)}\}_{i=1}^{N}$, inter-variable layer $f_v$, aggregation functions $\phi,\rho$
\STATE \textbf{Output:} Current-point system state $\mathbf{c}$ and anomaly score vector $\mathbf{a}$
\STATE Initialize token list $\mathcal{S}\leftarrow [\,]$
\FOR{$i=1$ to $N$}
    \STATE Extract variable window $\mathbf{X}^{(i)}\leftarrow \mathbf{X}_{[:,i,:]}\in\mathbb{R}^{B\times L}$
    \STATE Reshape $\mathbf{X}^{(i)}$ to $B\times L\times 1$
    \STATE Project scalar values to embeddings $\mathbf{E}^{(i)}\leftarrow \mathrm{Linear}(\mathbf{X}^{(i)})\in\mathbb{R}^{B\times L\times d}$
    \STATE Initialize the current anchor $\mathbf{z}^{(i)}\leftarrow \mathbf{E}^{(i)}_{[:,L,:]}\in\mathbb{R}^{B\times d}$
    \STATE Build current-relative lag indices $[L-1,\ldots,0]$ and look up lag embeddings $\mathbf{R}\in\mathbb{R}^{B\times L\times d_t}$
    \STATE Form key-value tokens $\mathbf{U}^{(i)}\leftarrow [\mathbf{E}^{(i)},\mathbf{R}]\in\mathbb{R}^{B\times L\times(d+d_t)}$
    \FOR{Temporal block $m=1$ to $M$}
        \STATE $\mathbf{q}\leftarrow \mathrm{LN}_q(\mathbf{z}^{(i)})\mathbf{W}_Q\in\mathbb{R}^{B\times 1\times d}$
        \STATE $\mathbf{K}\leftarrow \mathrm{LN}_{kv}(\mathbf{U}^{(i)})\mathbf{W}_K$, \quad $\mathbf{V}\leftarrow \mathrm{LN}_{kv}(\mathbf{U}^{(i)})\mathbf{W}_V$
        \STATE Split $\mathbf{q},\mathbf{K},\mathbf{V}$ into $H$ heads
        \STATE $\boldsymbol{\alpha}\leftarrow \mathrm{softmax}(\mathbf{q}\mathbf{K}^{\top}/\sqrt{d/H})$
        \STATE $\mathbf{h}\leftarrow \mathrm{merge}(\boldsymbol{\alpha}\mathbf{V})\mathbf{W}_O\in\mathbb{R}^{B\times 1\times d}$
        \STATE $\mathbf{g}\leftarrow \sigma(\mathrm{Linear}(\mathbf{q}))$
        \STATE $\tilde{\mathbf{z}}\leftarrow \mathbf{g}\odot\mathbf{q}+(1-\mathbf{g})\odot\mathbf{h}$
        \STATE $\mathbf{z}^{(i)}\leftarrow \mathrm{squeeze}(\tilde{\mathbf{z}})+\mathrm{FFN}(\mathrm{LN}(\mathrm{squeeze}(\tilde{\mathbf{z}})))$
    \ENDFOR
    \STATE Append $\mathrm{LN}(\mathbf{z}^{(i)})$ to $\mathcal{S}$
\ENDFOR
\STATE Stack variable tokens $\mathbf{S}_0\leftarrow \mathrm{stack}(\mathcal{S},\mathrm{dim}=1)\in\mathbb{R}^{B\times N\times d}$
\STATE Contextualize and normalize variables $\mathbf{S}\leftarrow f_v(\mathbf{S}_0)\in\mathbb{R}^{B\times N\times d}$
\STATE Pool variable states $\mathbf{c}\leftarrow \rho(\sum_{i=1}^{N}\phi(\mathbf{s}^{(i)}))\in\mathbb{R}^{B\times h}$
\STATE Compute fixed standard-normal NLL $a_b\leftarrow \frac{1}{2}\sum_{j=1}^{h}\left(c_{b,j}^{2}+\log(2\pi)\right)$ for each sample $b$
\STATE \textbf{return} $\mathbf{c},\mathbf{a}$
\end{algorithmic}
\end{algorithm}

Training and inference use the same forward path in Algorithm~\ref{alg:tenet_forward}.
During training, Algorithm~\ref{alg:tenet_forward} is applied to normal windows and the model minimizes $\mathcal{L}_{\mathrm{NLL}}=B^{-1}\sum_{b=1}^{B}a_b$ with AdamW.
During inference, the returned NLL value $a_b$ is used directly as the raw anomaly score for the current time point.
The raw scores are evaluated with AUC-ROC, AUC-PR, VUS-ROC, and VUS-PR without score smoothing, point adjustment, or reconstruction discrepancy.

\newpage
\section{Datasets}
\label{appendix:datasets}

\begin{table}[H]
\centering
\scriptsize
\setlength{\tabcolsep}{5pt}
\renewcommand{\arraystretch}{1.3}
\caption{Statistics of the benchmark datasets used in the experiments. Anomaly count, anomaly ratio, and segment lengths are computed from the test labels, where count denotes the number of contiguous anomalous segments.}
\label{tab:appendix-datasets}
\begin{tabular}{c|c|c|c|c}
\toprule
\textbf{Dataset}& \textbf{PSM} & \textbf{SMD} & \textbf{SWaT} & \textbf{SWAN} \\
\midrule
\textbf{Features} & 25 & 38 & 51 & 38 \\
\midrule

\multirow{2}{*}{\textbf{Set size}} &
Train: 132{,}481 & Train: 23{,}697  & Train: 496{,}800 & Train: 60{,}000 \\
 & Test: 87{,}841  & Test: 23{,}697  & Test: 449{,}919 & Test: 60{,}000 \\
\midrule

\multirow{2}{*}{\textbf{Anomalies}} &
Count: 71 & Count: 13 & Count: 34 & Count: 6{,}148 \\
 & Ratio: 27.73\% & Ratio: 10.12\% & Ratio: 12.02\% & Ratio: 32.60\% \\
\midrule

\multirow{3}{*}{\textbf{Segment Lengths}} &
Min: 1 & Min: 3 & Min: 101 & Min: 1 \\
 & Med: 5 & Med: 26 & Med: 447 & Med: 1 \\
 & Max: 8{,}861 & Max: 1{,}215 & Max: 35{,}900 & Max: 11{,}442 \\
\midrule

\textbf{Sampling Period} & 1 min & 1 min & 1 sec & 12 min \\
\bottomrule
\end{tabular}
\end{table}

\subsection{Dataset Selection}
\label{appendix:dataset-selection}

For the main benchmark, we considered real-world datasets with at least 20 variables as a practical scope for studying variable-wise temporal encoding and cross-variable modeling.
We selected complementary anomaly durations and frequencies rather than aiming to cover every domain.
PSM includes short transients and sustained anomalies, with labeled segment lengths from 1 to 8{,}861 points.
SMD (\texttt{machine-1-7}) tests detection with a lower anomaly ratio of 10.12\%, while SWaT includes sustained process anomalies lasting up to 35{,}900 points.
SWAN tests point-like events, with 6{,}148 labeled segments and a median length of one point.
We use the original releases to preserve their train--test splits and sequence boundaries, rather than selecting these four datasets from TAB's curated collection.

\subsection{Dataset Statistics}
\label{appendix:datasets-statistics}

Table~\ref{tab:appendix-datasets} summarizes the benchmark datasets used in our experiments.
We report not only the dimensionality and split sizes, but also the segment-level characteristics
of anomalies.
Here, count denotes the number of contiguous anomalous segments in the test labels,
and segment lengths are computed over these contiguous segments (minimum/median/maximum).
These datasets differ substantially in terms of dimensionality, time series length, anomaly density,
and the temporal structure of anomalous events.

\textbf{PSM}~\cite{PSM-2021} contains relatively few but potentially long anomaly segments, resulting in a high anomaly
ratio despite a limited number of events.
\textbf{SMD}~\cite{OmniAnomaly-2019} (\texttt{machine-1-7}) has a moderate number of short-to-medium anomaly segments with ${\sim}10\%$ density.
\textbf{SWaT}~\cite{SWaT-2016} exhibits a wide range of anomaly durations, spanning from short abnormal intervals
to extremely long segments, which stresses robustness under heterogeneous attack effects.
\textbf{SWAN}~\cite{SWAN-2020} has the densest anomaly structure among our benchmarks,
containing thousands of very short anomaly segments intermixed with longer ones,
which tests stability under heavily contaminated streams.

\subsection{Dataset Descriptions}
\label{appendix:datasets-descriptions}

\paragraph{PSM (Pooled Server Metrics).}
PSM~\cite{PSM-2021} is a multivariate server-metrics dataset released in the context of practical anomaly detection
research, containing 25 variables with minute-level sampling.
Its test split includes both short spikes and prolonged deviations, creating a challenging mixture
of dense and long-duration anomalies.

\paragraph{SMD (Server Machine Dataset).}
SMD~\cite{OmniAnomaly-2019} consists of 28 independent machine subsets, each with 38 variables.
We report results on \texttt{machine-1-7}, following recent evaluation practice that avoids averaging over highly heterogeneous machines.
This machine contains 23{,}697 training samples and 23{,}697 test samples at 1-minute sampling.

\paragraph{SWaT (Secure Water Treatment).}
SWaT~\cite{SWaT-2016} is an industrial control system (ICS) dataset collected from a realistic water treatment testbed,
with 51 tags (sensors and actuators) sampled every second.
It was recorded over 11 days of continuous operation, where the first 7 days correspond to normal operation
and the remaining days include cyber/physical attack interventions.
The test split contains attack periods that induce diverse anomaly durations and multivariate dependency shifts.

\paragraph{SWAN (Space Weather ANalytics).}
SWAN~\cite{SWAN-2020} is a solar-activity monitoring dataset collected from NASA SDO/HMI vector magnetograms,
with 38 numerical features describing physical properties of active regions (e.g., total unsigned flux, magnetic free energy, helicity).
Samples are recorded every 12 minutes, and anomalies correspond to pre-flare and flare states of solar active regions.
The train and test splits each contain 60{,}000 time steps, and anomalies appear as densely packed short segments mixed with longer event windows.

\subsection{Dataset Sources and Access}
\label{appendix:datasets-sources}

We use publicly available benchmarks under their respective terms of use:

\begin{itemize}
    \item \textbf{PSM (Pooled Server Metrics)}\footnote
    {\href{https://github.com/eBay/RANSynCoders}
    {https://github.com/eBay/RANSynCoders}}:
    Released as part of the \texttt{eBay/RANSynCoders} repository.
    The repository documentation specifies that the sample data are released under Creative Commons Attribution 4.0
    (see \texttt{data/LICENSE}), while the non-data repository content is released under BSD-3 (see \texttt{LICENSE}).

    \item \textbf{SMD (Server Machine Dataset)}\footnote
    {\href{https://github.com/NetManAIOps/OmniAnomaly}
    {https://github.com/NetManAIOps/OmniAnomaly}}:
    Distributed via the \texttt{NetManAIOps/OmniAnomaly} repository, which states that the 28 subsets should be trained and tested separately, and provides an MIT license for the repository.

    \item \textbf{SWaT (Secure Water Treatment)}\footnote
    {\href{https://itrust.sutd.edu.sg/itrust-labs_datasets/dataset_info}
    {https://itrust.sutd.edu.sg/itrust-labs\_datasets/dataset\_info/}}:
    Provided by iTrust, SUTD. Access typically requires a request via the dataset portal,
    and users must comply with the stated terms (e.g., attribution requirements and restrictions on redistribution).

    \item \textbf{SWAN (Space Weather ANalytics)}\footnote
    {\href{https://dataverse.harvard.edu/dataset.xhtml?persistentId=doi:10.7910/DVN/EBCFKM}
    {https://dataverse.harvard.edu/dataset.xhtml?persistentId=doi:10.7910/DVN/EBCFKM}}:
    Released on Harvard Dataverse under an open license for academic research on solar activity and space-weather forecasting.
\end{itemize}

\newpage
\section{Evaluation Metrics}
\label{appendix:metrics}
\begin{table}[H]
\centering
\scriptsize
\setlength{\tabcolsep}{4pt}
\renewcommand{\arraystretch}{1.15}
\caption{Summary of the evaluation metrics used in the experiments.}
\label{tab:appendix-metrics}
\begin{tabular}{c|c|c|c|c}
\toprule
 \textbf{Metric}& \textbf{AUC-ROC} & \textbf{AUC-PR} & \textbf{VUS-ROC} & \textbf{VUS-PR} \\
\midrule
\textbf{Granularity}        & Point-wise & Point-wise & Point+Time & Point+Time \\
\midrule
\textbf{Threshold-Free}     & Yes        & Yes        & Yes        & Yes \\
\midrule
\textbf{Temporal-Aware}     & No         & No         & Yes        & Yes \\
\midrule
\textbf{Delay-Tolerant}     & No         & No         & Yes        & Yes \\
\midrule
\textbf{Imbalance-Robust}   & No         & Yes        & Yes        & Yes \\
\bottomrule
\end{tabular}
\end{table}

We evaluate detection performance using four threshold-free metrics, \textbf{AUC-ROC}~\cite{ROC-2006}, \textbf{AUC-PR}~\cite{PR_ROC-2006}, \textbf{VUS-ROC}, and \textbf{VUS-PR}~\cite{VUS-2022}.
These metrics assess the ranking quality of anomaly scores without committing to a fixed operating threshold.
Since time-series anomalies are often sparse and temporally structured, we report both ROC and PR metrics and additionally include range-aware VUS metrics to account for temporal tolerance around anomalous intervals.

Unless stated otherwise, we directly use the raw anomaly scores produced by the model, without any score post-processing.
We focus on threshold-free metrics because evaluating a detector with a test-set optimal threshold is not operationally meaningful.
It implicitly assumes access to future test labels and overestimates deploy-time performance.
Using a pre-selected threshold is also method-dependent and typically requires strong assumptions (e.g., consistent train/test score distributions and stable score distributions across deployment data),
which may not hold in real systems.

\subsection{AUC-ROC and AUC-PR}
\label{sec:auc}

Let $\{s_t\}_{t=1}^{T}$ be the anomaly score sequence and $\{y_t\}_{t=1}^{T}$ the binary labels.
For a threshold $\tau$, predictions are $\hat{y}_t(\tau)=\mathbb{I}[s_t \ge \tau]$, yielding
$TP(\tau)$, $FP(\tau)$, $TN(\tau)$, and $FN(\tau)$.
The true positive rate (TPR) and false positive rate (FPR) are
\begin{equation}
\begin{aligned}
\mathrm{TPR}(\tau) &= \frac{TP(\tau)}{TP(\tau)+FN(\tau)},\\
\mathrm{FPR}(\tau) &= \frac{FP(\tau)}{FP(\tau)+TN(\tau)}.
\end{aligned}
\label{eq:tpr_fpr}
\end{equation}
AUC-ROC is defined as the area under the ROC curve parameterized by $\tau$:
\begin{equation}
\mathrm{AUC\text{-}ROC}
=
\int \mathrm{TPR}(\tau)\, d\,\mathrm{FPR}(\tau).
\label{eq:auc_roc}
\end{equation}
In practice, these AUC metrics are computed by sweeping $\tau$ over all distinct score values (i.e., all possible decision thresholds), as implemented in standard toolkits.

Similarly, precision and recall at threshold $\tau$ are
\begin{equation}
\begin{aligned}
\mathrm{Precision}(\tau) &= \frac{TP(\tau)}{TP(\tau)+FP(\tau)},\\
\mathrm{Recall}(\tau) &= \frac{TP(\tau)}{TP(\tau)+FN(\tau)}.
\end{aligned}
\label{eq:prec_rec}
\end{equation}
AUC-PR is the area under the precision--recall curve:
\begin{equation}
\mathrm{AUC\text{-}PR}
=
\int \mathrm{Precision}(\tau)\, d\,\mathrm{Recall}(\tau).
\label{eq:auc_pr}
\end{equation}

In multivariate time-series anomaly detection, anomalies are typically rare.
As a result, AUC-ROC can be overly optimistic because $\mathrm{FPR}$ is diluted by the large number of
true negatives.
AUC-PR is often more informative in this regime since it focuses on the positive class and better reflects
the practical trade-off between false alarms and missed anomalies.
We therefore report both metrics, and interpret AUC-PR as the more reliable indicator under severe imbalance.
Both AUC-ROC and AUC-PR are temporally agnostic.
They do not measure detection continuity, localization lag,
or tolerance around anomalous ranges.
To explicitly account for temporal tolerance while remaining threshold-free, we additionally report VUS metrics.

\subsection{VUS-ROC and VUS-PR}
\label{sec:vus}

Our implementation of Volume Under the Surface (VUS) metrics follows a range-aware threshold-free evaluation scheme.
Let $\Omega=\{0,1,\dots,10\}$ be the set of temporal tolerance windows used in the main benchmark.
For each $\omega\in\Omega$, we construct a softened label sequence $\tilde{y}^{(\omega)}_t\in[0,1]$ by extending each anomalous interval by $\lfloor \omega/2 \rfloor$ time steps on both sides and assigning distance-dependent fractional credit near the interval boundaries.
Given anomaly scores $\{s_t\}_{t=1}^{T}$ and threshold $\tau$, predictions are $\hat{y}_t(\tau)=\mathbb{I}[s_t\ge\tau]$.
For each $\omega$ and $\tau$, we compute soft true-positive, false-positive, false-negative, and true-negative counts as
\begin{equation}
\begin{aligned}
\mathrm{TP}_{\omega}(\tau)&=\sum_t \tilde{y}^{(\omega)}_t \hat{y}_t(\tau),
&
\mathrm{FP}_{\omega}(\tau)&=\sum_t \left(1-\tilde{y}^{(\omega)}_t\right)\hat{y}_t(\tau),\\
\mathrm{FN}_{\omega}(\tau)&=\sum_t \tilde{y}^{(\omega)}_t-\mathrm{TP}_{\omega}(\tau),
&
\mathrm{TN}_{\omega}(\tau)&=\sum_t \left(1-\tilde{y}^{(\omega)}_t\right)-\mathrm{FP}_{\omega}(\tau).
\end{aligned}
\end{equation}
These quantities define tolerance-dependent ROC and PR surfaces through $\mathrm{TPR}_{\omega}(\tau)$, $\mathrm{FPR}_{\omega}(\tau)$, and $\mathrm{Precision}_{\omega}(\tau)$.
VUS is then computed by averaging the ROC/PR surfaces over $\omega$ and integrating the resulting mean curves:
\begin{equation}
\begin{aligned}
\mathrm{VUS\text{-}ROC}
&=
\int \overline{\mathrm{TPR}}(\tau)\, d\,\overline{\mathrm{FPR}}(\tau),\\
\mathrm{VUS\text{-}PR}
&=
\int \overline{\mathrm{Precision}}(\tau)\, d\,\overline{\mathrm{TPR}}(\tau),
\end{aligned}
\label{eq:vus}
\end{equation}
where the overline denotes averaging over $\omega\in\Omega$.
In the implementation, $\tau$ is swept over 200 uniformly spaced thresholds between the minimum and maximum anomaly scores.

VUS therefore measures how consistently a method ranks anomalies highly under varying temporal tolerance around anomalous ranges.
In our implementation, it is a range-aware extension of ROC/PR evaluation rather than an event-counting metric.
We report both VUS-ROC and VUS-PR to capture tolerance-aware performance under class imbalance, and we keep the
metric implementations fixed across all methods for fair comparison.

\newpage
\section{Baselines}
\label{appendix:baselines}

\subsection{Evaluation Protocol}
\label{appendix:baselines-protocol}

For the main benchmark in Table~\ref{tab:main}, methods share data loading, normalization, metric computation, and evaluation procedures.
All reported baseline values are obtained from our own reruns, not copied from TAB's public leaderboard.
TAB provides model implementations and pipeline components, but our configuration is not an exact reproduction of its leaderboard protocol.
Each baseline follows the hyperparameters recommended in its original paper or official implementation, including its window length and method-specific training duration.
The 10-epoch setting in Section~\ref{sec:implementation} applies only to TENET and is not imposed on the baselines.
We use seeds $\{10,23,30,31,37\}$ for the five-run comparisons, and we do not apply any additional score post-processing
(e.g., smoothing or calibration) beyond each method's original score.

For fair comparison, we standardize the evaluation interface across methods.
Each baseline produces a scalar anomaly score per time step, and all scores are evaluated by the same metric code.
When a model naturally outputs window-level quantities (e.g., reconstructed windows, predicted values, likelihoods, or
attention-based scores), we reproduce the anomaly scoring rule of the original method and consistently map it to a
point-wise score at the current target time step.
Input normalization is performed with \texttt{StandardScaler} fitted on the training split and applied to both training
and test splits.

Our shared settings differ from TAB's public configuration in four respects.
Models requiring validation use a chronological 90/10 train--validation split rather than 80/20.
We retain the original dataset releases and keep independent sequences separate rather than concatenating them.
Train-fitted standardization is applied to classical as well as deep baselines.
For the main benchmark, the VUS tolerance range is fixed to $\Omega=\{0,\ldots,10\}$ rather than selected separately for each dataset.

Most trainable baselines are evaluated with five independent random seeds and we report seed-averaged performance.
For foundation-style models that provide official pre-trained checkpoints, we run the released pre-trained weights and
follow the authors' prescribed anomaly detection recipe, rather than re-pretraining from scratch.
Overall, model-specific differences are preserved (including window length and scoring rule), while the evaluation code
path and metric computation remain identical across methods.

\subsection{Code Resources}
\label{appendix:baselines-repos}

We evaluate most baselines through the unified \textbf{TAB benchmark} codebase\footnote
{\href{https://github.com/decisionintelligence/TAB}
{https://github.com/decisionintelligence/TAB}}.
For methods that require dedicated code paths, we use their official implementations.
\begin{itemize}
    \setlength{\itemsep}{0.0em}
    \setlength{\parsep}{0pt}
    \item \textbf{OmniAnomaly} (KDD 2019)\footnote
    {\href{https://github.com/NetManAIOps/OmniAnomaly}
    {https://github.com/NetManAIOps/OmniAnomaly}}
    \item \textbf{SARAD} (NeurIPS 2024)\footnote
    {\href{https://github.com/daidahao/SARAD}
    {https://github.com/daidahao/SARAD}}
    \item \textbf{Chronos} (TMLR 2024)\footnote
    {\href{https://github.com/amazon-science/chronos-forecasting}
    {https://github.com/amazon-science/chronos-forecasting}}
    \item \textbf{CAROTS} (ICML 2025)\footnote
    {\href{https://github.com/kimanki/CAROTS}
    {https://github.com/kimanki/CAROTS}}
\end{itemize}




\newpage
\section{System Configuration}
\label{appendix:system-configuration}

All TENET experiments are conducted using the following system configuration.
\begin{itemize}
    \setlength{\itemsep}{0.0em}
    \setlength{\parsep}{0pt}
    \item \textbf{OS:} Ubuntu 24.04.3 LTS.
    \item \textbf{Python:} Python 3.10.19.
    \item \textbf{GPU:} NVIDIA RTX PRO 6000 Blackwell Workstation Edition with 96 GB GPU memory.
    \item \textbf{CPU:} AMD Ryzen 9 9950X3D 16-core processor with 32 hardware threads.
    \item \textbf{System memory:} 91 GiB.
    \item \textbf{CUDA:} CUDA toolkit 12.8.1 and CUDA runtime 12.8.90.
    \item \textbf{PyTorch:} PyTorch 2.11.0+cu128.
\end{itemize}

\newpage
\section{Statistical Significance}
\label{appendix:statistical_significance}
\begin{table}[H]
\caption{Standard deviation of anomaly detection performance on four multivariate time series anomaly detection datasets. All metrics are reported in percent and computed over five random seeds. Lower values indicate higher stability across runs.}
\label{tab:statistical_significance}
\centering
\setlength{\tabcolsep}{1.5pt}
\renewcommand{\arraystretch}{0.95}
\begin{scriptsize}
\begin{tabular}{c|cccc|cccc|cccc|cccc}
\toprule
\multicolumn{1}{c|}{\textbf{Dataset}} &
\multicolumn{4}{c|}{\textbf{PSM (25 vars)}} &
\multicolumn{4}{c|}{\textbf{SMD (38 vars)}} &
\multicolumn{4}{c|}{\textbf{SWaT (51 vars)}} &
\multicolumn{4}{c}{\textbf{SWAN (38 vars)}} \\
\midrule
\textbf{Method} &
\textbf{A-ROC} & \textbf{A-PR} & \textbf{V-ROC} & \textbf{V-PR} &
\textbf{A-ROC} & \textbf{A-PR} & \textbf{V-ROC} & \textbf{V-PR} &
\textbf{A-ROC} & \textbf{A-PR} & \textbf{V-ROC} & \textbf{V-PR} &
\textbf{A-ROC} & \textbf{A-PR} & \textbf{V-ROC} & \textbf{V-PR} \\
\midrule

IF
& -- & -- & -- & --
& -- & -- & -- & --
& -- & -- & -- & --
& -- & -- & -- & -- \\
PCA
& -- & -- & -- & --
& -- & -- & -- & --
& -- & -- & -- & --
& -- & -- & -- & -- \\
OCSVM
& -- & -- & -- & --
& -- & -- & -- & --
& -- & -- & -- & --
& -- & -- & -- & -- \\
\midrule
DAGMM
& 2.86 & 2.05 & 2.89 & 2.77
& 1.96 & 3.07 & 2.02 & 4.83
& 6.29 & 2.35 & 19.03 & 15.14
& 4.66 & 7.57 & 2.91 & 5.00 \\
THOC
& 3.77 & 5.00 & 3.73 & 4.97
& 3.77 & 4.58 & 10.56 & 9.99
& 0.95 & 1.66 & 15.79 & 13.57
& 3.50 & 5.78 & 3.39 & 3.56 \\
\midrule
DC
& 0.29 & 0.24 & 0.56 & 0.82
& 0.11 & 0.03 & 0.03 & 0.25
& 0.29 & 0.03 & 0.10 & 0.31
& 0.18 & 0.08 & 0.03 & 0.71 \\
CAROTS
& 1.21 & 1.96 & 2.31 & 0.07
& 2.09 & 5.56 & 2.52 & 4.80
& 1.63 & 2.21 & 1.30 & 1.57
& 5.80 & 4.60 & 2.20 & 2.50 \\
\midrule

Omni
& 1.89 & 2.34 & 0.56 & 0.52
& 0.11 & 1.37 & 7.75 & 8.23
& 0.07 & 0.02 & 0.11 & 3.58
& 0.78 & 0.72 & 0.96 & 0.55 \\
A.T
& 2.10 & 0.88 & 0.95 & 13.09
& 1.20 & 0.18 & 0.16 & 13.55
& 4.39 & 0.41 & 0.12 & 0.42
& 4.19 & 1.13 & 0.07 & 5.30 \\
PatchTST
& 0.06 & 0.10 & 0.05 & 0.76
& 0.33 & 0.11 & 0.16 & 1.09
& 0.07 & 0.02 & 0.02 & 0.03
& 0.46 & 0.38 & 0.22 & 0.36 \\
TimesNet
& 0.56 & 0.46 & 0.03 & 1.40
& 0.33 & 0.27 & 0.16 & 1.36
& 0.32 & 0.09 & 0.10 & 0.76
& 0.36 & 0.61 & 0.07 & 1.20 \\
iTrans
& 0.11 & 0.11 & 0.24 & 1.02
& 0.30 & 0.13 & 0.04 & 1.06
& 0.16 & 0.01 & 0.08 & 0.20
& 0.29 & 0.32 & 0.26 & 0.58 \\
Modern
& 0.14 & 0.18 & 0.03 & 0.46
& 0.80 & 0.39 & 0.08 & 9.74
& 0.08 & 0.18 & 0.08 & 0.19
& 0.19 & 0.33 & 0.07 & 0.19 \\
SARAD
& 1.52 & 1.26 & 4.98 & 1.75
& 0.42 & 0.69 & 0.03 & 0.12
& 0.91 & 12.61 & 0.56 & 9.02
& 1.09 & 0.90 & 0.07 & 0.09 \\
CATCH
& 0.78 & 0.75 & 5.14 & 0.87
& 0.18 & 0.15 & 0.03 & 1.60
& 0.27 & 0.35 & 0.16 & 0.73
& 0.08 & 0.10 & 0.06 & 0.22 \\
CrossAD
& 1.30 & 0.98 & 0.01 & 0.66
& 0.04 & 0.50 & 0.01 & 0.07
& 3.42 & 0.91 & 0.11 & 0.24
& 0.12 & 0.10 & 0.03 & 0.04 \\
\midrule

TTM
& 0.06 & 0.13 & 0.02 & 0.69
& 0.41 & 0.15 & 0.00 & 0.00
& 0.01 & 0.05 & 0.07 & 0.11
& 0.19 & 0.16 & 0.01 & 0.03 \\
Timer
& 0.02 & 0.01 & 0.02 & 0.05
& 0.09 & 0.32 & 0.00 & 0.46
& 0.01 & 0.01 & 0.05 & 0.06
& 0.19 & 0.08 & 0.01 & 0.01 \\
Chronos
& -- & -- & -- & --
& -- & -- & -- & --
& -- & -- & -- & --
& -- & -- & -- & -- \\
GPT4TS
& 0.36 & 0.51 & 0.07 & 0.55
& 0.13 & 0.15 & 0.00 & 0.35
& 0.42 & 0.05 & 0.02 & 2.75
& 1.57 & 1.96 & 0.08 & 0.84 \\
\midrule

\textbf{TENET}
& 0.69 & 1.73 & 0.68 & 1.70
& 1.19 & 1.63 & 1.19 & 1.63
& 0.47 & 0.96 & 0.48 & 0.96
& 0.96 & 1.01 & 0.94 & 0.72 \\
\bottomrule
\end{tabular}
\end{scriptsize}
\end{table}

To assess robustness to random initialization and stochastic optimization, we report the standard deviation of threshold-free ranking metrics across independent runs.
Table~\ref{tab:statistical_significance} summarizes the run-to-run variability of AUC-ROC (A-ROC), AUC-PR (A-PR), VUS-ROC (V-ROC), and VUS-PR (V-PR), computed over five random seeds for each trainable method.
Lower values indicate higher stability across runs.
For TTM, Timer, and GPT4TS, we initialize from released pre-trained weights and fine-tune the models on each dataset before evaluation.
Chronos is evaluated using released pre-trained weights, but we do not report standard deviations because repeated runs were not conducted for this model.
Dashes indicate cases where repeated-run standard deviations are unavailable.

\newpage
\section{One-Class Objective Configurations}
\label{appendix:oneclass_objectives}

This appendix specifies the implementation details of the one-class objectives used in the encoder--objective comparisons in Table~\ref{tab:encoder_framework} and Table~\ref{tab:encoder_framework_ext}.
For a fixed encoder, all four objectives receive the same pooled current-point state $\mathbf{c}\in\mathbb{R}^{h}$ with $h=256$.
Objective-specific scoring modules and losses differ, while the optimizer (AdamW), batch size, learning rate, and number of epochs follow the corresponding NLL experiment for the same encoder and dataset.
This keeps the encoder interface and main training protocol comparable across one-class objectives.

\paragraph{NLL (standard normal reference).}
The NLL head is parameter-free.
For each pooled state $\mathbf{c}$, the per-sample loss and anomaly score are
\[
\mathcal{L}_{\mathrm{NLL}}(\mathbf{c})
=
\tfrac{1}{2}\bigl(\|\mathbf{c}\|_2^{2} + h\log 2\pi\bigr),
\qquad
s_{\mathrm{NLL}}(\mathbf{c}) = \tfrac{1}{2}\bigl(\|\mathbf{c}\|_2^{2} + h\log 2\pi\bigr).
\]
Training minimizes the mean of $\mathcal{L}_{\mathrm{NLL}}$ over normal training windows, which corresponds to fitting the pooled state to a fixed standard normal reference $\mathcal{N}(\mathbf{0},\mathbf{I}_h)$ without learning any additional parameters.

\paragraph{DeepSVDD (fixed-center one-class).}
A center vector $\boldsymbol{\mu}\in\mathbb{R}^{h}$ is computed once, before training, by averaging encoder outputs over the first 50 training batches with the encoder in evaluation mode.
Following the original protocol of \citet{DeepSVDD-2018}, components of $\boldsymbol{\mu}$ with $|\mu_i|<0.01$ are clamped to $0.01$ to avoid the trivial solution at the origin.
$\boldsymbol{\mu}$ is then registered as a fixed buffer.
For each pooled state, the loss and score are
\[
\mathcal{L}_{\mathrm{SVDD}}(\mathbf{c})
=
\|\mathbf{c}-\boldsymbol{\mu}\|_2^{2},
\qquad
s_{\mathrm{SVDD}}(\mathbf{c})=\|\mathbf{c}-\boldsymbol{\mu}\|_2^{2}.
\]
Only the encoder parameters are updated during training, and $\boldsymbol{\mu}$ is not updated.

\paragraph{THOC.}
THOC uses multiple hyperspheres and hypersphere centers to define a one-class objective~\citep{THOC-2020}.
In our implementation, a set of $K=8$ learnable hypersphere centers $\mathbf{P}\in\mathbb{R}^{K\times h}$ is maintained on the unit sphere by re-normalizing each center to unit $\ell_2$ norm at every forward pass.
For pooled state $\mathbf{c}$, we compute soft assignments to the centers and a weighted cosine distance:
\[
\mathbf{R}=\mathrm{softmax}\bigl(\hat{\mathbf{c}}\,\mathbf{P}^{\top}\bigr)\in\mathbb{R}^{K},
\qquad
s_{\mathrm{THOC}}(\mathbf{c})
=
\sum_{k=1}^{K} R_{k}\bigl(1-\cos(\mathbf{c},\mathbf{P}_{k})\bigr),
\]
where $\hat{\mathbf{c}}=\mathbf{c}/(\|\mathbf{c}\|_2+\epsilon)$, with a small $\epsilon$ for numerical stability.
The training loss combines the mean of $s_{\mathrm{THOC}}$ with an orthogonality regularizer on the hypersphere centers:
\[
\mathcal{L}_{\mathrm{THOC}}
=
\frac{1}{B}\sum_{b=1}^{B} s_{\mathrm{THOC}}(\mathbf{c}_b)
+ \lambda_{\mathrm{orth}}\,\bigl\|\mathbf{P}\mathbf{P}^{\top}-\mathbf{I}_{K}\bigr\|_{F}^{2}.
\]
We set $\lambda_{\mathrm{orth}}=1.0$ in all objective-comparison experiments.
Both encoder and hypersphere-center parameters are jointly optimized.

\paragraph{DAGMM (GMM-density one-class).}
A compression network $g_{\psi}\colon\mathbb{R}^{h}\!\to\!\mathbb{R}^{16}$ produces a low-dimensional latent $\mathbf{z}_{c}=g_{\psi}(\mathbf{c})$.
We then form an extended representation $\mathbf{z}\in\mathbb{R}^{18}$ by appending $\|\mathbf{c}\|_2$ and $\|\mathbf{z}_{c}\|_2$ to $\mathbf{z}_{c}$, mirroring the reconstruction-based norm features used by \citet{DAGMM-2018}.
An estimation network with a softmax output predicts soft GMM assignments $\boldsymbol{\gamma}\in\mathbb{R}^{n_{\mathrm{gmm}}}$ for $n_{\mathrm{gmm}}=4$ components.
GMM parameters $(\boldsymbol{\pi},\boldsymbol{\mu}_{1{:}n_{\mathrm{gmm}}},\boldsymbol{\Sigma}_{1{:}n_{\mathrm{gmm}}})$ are stored as buffers and updated at training time using an exponential moving average with factor $0.01$, computed from the detached $(\mathbf{z},\boldsymbol{\gamma})$ to avoid backpropagating through the GMM update.
The per-sample training loss and inference score are the negative log-likelihood under the diagonal-covariance GMM,
\[
\mathcal{L}_{\mathrm{DAGMM}}(\mathbf{c})
=
s_{\mathrm{DAGMM}}(\mathbf{c})
=
-\log\sum_{k=1}^{n_{\mathrm{gmm}}}\pi_{k}\,\mathcal{N}\!\bigl(\mathbf{z}\mid\boldsymbol{\mu}_{k},\boldsymbol{\Sigma}_{k}\bigr),
\]
where each $\boldsymbol{\Sigma}_{k}$ is regularized with a diagonal $10^{-3}\mathbf{I}$ for numerical stability.
Encoder, compression, and estimation parameters are jointly optimized, while GMM buffers are updated via the EMA rule.

\newpage
\section{Additional Experiments}
\label{appendix:additional-experiments}

\subsection{Additional Encoder--Objective Comparison}
\label{appendix:encoder_ext}

\begin{table}[H]
\caption{Encoder and one-class objective comparison on SWAN. All metrics are reported in percent and averaged over five random seeds. \textbf{Bold} indicates the best score.}
\label{tab:encoder_framework_ext}
\centering
\setlength{\tabcolsep}{1.3pt}
\renewcommand{\arraystretch}{1.05}
\begin{scriptsize}
\begin{tabular}{c|cccc|cccc|cccc|cccc}
\toprule
\textbf{Objective}
& \multicolumn{4}{c|}{\textbf{NLL}} & \multicolumn{4}{c|}{\textbf{DeepSVDD}} & \multicolumn{4}{c|}{\textbf{THOC}} & \multicolumn{4}{c}{\textbf{DAGMM}} \\
\midrule
\textbf{Metric}
& \textbf{A-ROC} & \textbf{A-PR} & \textbf{V-ROC} & \textbf{V-PR}
& \textbf{A-ROC} & \textbf{A-PR} & \textbf{V-ROC} & \textbf{V-PR}
& \textbf{A-ROC} & \textbf{A-PR} & \textbf{V-ROC} & \textbf{V-PR}
& \textbf{A-ROC} & \textbf{A-PR} & \textbf{V-ROC} & \textbf{V-PR} \\
\midrule
LSTM
& 73.71 & 64.50 & 64.30 & 72.30
& 78.99 & 70.41 & 68.64 & 76.06
& 58.26 & 47.26 & 55.39 & 63.39
& 75.11 & 68.32 & 59.33 & 79.04 \\
Dilated RNN
& 68.00 & 57.53 & 62.11 & 69.24
& 79.59 & 72.75 & 67.87 & 76.25
& 64.48 & 55.54 & 60.81 & 67.86
& 78.21 & 71.73 & 61.55 & 78.05 \\
Mamba
& 52.75 & 36.66 & 51.85 & 57.72
& 82.70 & 66.93 & 70.17 & 73.81
& 75.18 & 57.75 & 64.85 & 68.22
& 68.81 & 51.94 & 62.21 & 66.58 \\
ModernTCN
& 63.38 & 54.01 & 58.21 & 73.65
& 63.37 & 53.97 & 58.14 & 73.71
& 63.40 & 47.90 & 61.43 & 66.45
& 62.01 & 52.19 & 57.56 & 72.38 \\
GPT4TS
& 80.46 & 71.79 & 66.61 & 77.42
& 75.89 & 65.20 & 61.87 & 73.64
& 78.68 & 69.41 & 57.97 & 78.78
& 47.93 & 39.02 & 48.36 & 58.76 \\
\midrule
\textbf{TENET}
& \textbf{88.36} & \textbf{84.07} & \textbf{72.98} & \textbf{80.97}
& \textbf{89.01} & \textbf{85.18} & \textbf{73.60} & \textbf{81.63}
& \textbf{88.67} & \textbf{84.46} & \textbf{73.51} & \textbf{81.35}
& \textbf{84.39} & \textbf{79.21} & \textbf{71.10} & \textbf{79.31} \\
\bottomrule
\end{tabular}
\end{scriptsize}
\end{table}

We further examine the encoder--objective trend by repeating the same comparison on SWAN.
The protocol, metrics, objectives, and pooled-state dimensionality are kept unchanged across encoders.
This setting examines whether TENET remains effective when the same representation is scored by different one-class objectives.

Table~\ref{tab:encoder_framework_ext} shows that TENET obtains the best score across all reported objective--metric pairs on SWAN.
On A-PR, TENET exceeds the next-best encoder by more than 13 percentage points under both NLL and DeepSVDD.
Together with the PSM results in Table~\ref{tab:encoder_framework}, this provides additional evidence that TENET's advantage is not specific to the default NLL score.
Rather, the results support our claim that constructing an informative current-point state remains important across different one-class objectives.

\subsection{Fixed Versus Learnable Gaussian References}
\label{appendix:gaussian-reference}

We also evaluate a learnable diagonal Gaussian reference on PSM and SMD, with score
\[
\mathcal{A}_{\mathrm{learn}}(\mathbf{c})
=\frac{1}{2}\sum_{j=1}^{h}
\left[\frac{(c_j-\mu_j)^2}{v_j}+\log v_j+\log(2\pi)\right].
\]
The means and variances are jointly optimized with TENET, initialized to zero and one, respectively.
We parameterize $v_j=\operatorname{softplus}(a_j)+10^{-6}$ to ensure positive variances.
The training settings and five seeds match the fixed-reference experiments.
The fixed-reference results in Table~\ref{tab:gaussian_reference} are taken from Table~\ref{tab:main}.

\begin{table}[H]
\caption{Gaussian reference comparison for \textbf{TENET} on PSM and SMD. Metrics are reported in percent as mean $\pm$ standard deviation over five seeds.}
\label{tab:gaussian_reference}
\centering
\setlength{\tabcolsep}{2pt}
\renewcommand{\arraystretch}{1.05}
\begin{scriptsize}
\resizebox{\linewidth}{!}{%
\begin{tabular}{c|cccc|cccc}
\toprule
\multirow{2}{*}{\textbf{Reference}} &
\multicolumn{4}{c|}{\textbf{PSM}} &
\multicolumn{4}{c}{\textbf{SMD}} \\
\cmidrule(lr){2-5}\cmidrule(lr){6-9}
& \textbf{A-ROC} & \textbf{A-PR} & \textbf{V-ROC} & \textbf{V-PR}
& \textbf{A-ROC} & \textbf{A-PR} & \textbf{V-ROC} & \textbf{V-PR} \\
\midrule
Fixed standard normal
& $86.14 \pm 0.69$ & $74.27 \pm 1.73$ & $85.82 \pm 0.68$ & $74.21 \pm 1.70$
& $90.04 \pm 1.19$ & $74.78 \pm 1.63$ & $89.50 \pm 1.19$ & $74.41 \pm 1.63$ \\
Learnable Gaussian
& $86.20 \pm 0.74$ & $74.44 \pm 1.68$ & $85.88 \pm 0.73$ & $74.38 \pm 1.65$
& $90.03 \pm 0.97$ & $74.77 \pm 1.27$ & $89.50 \pm 0.94$ & $74.42 \pm 1.20$ \\
\bottomrule
\end{tabular}
}
\end{scriptsize}
\end{table}

The reported differences are small relative to variation across runs and show no consistent advantage for the learnable reference.
We therefore retain the fixed standard normal reference for simplicity, while TENET also supports learning the reference parameters.

\newpage
\subsection{Cross-Model Computational Cost}
\label{appendix:cross-model-cost}

Table~\ref{tab:cross_model_cost} compares representative baselines with TENET on synthetic inputs with $N=L=100$, using FP32 precision on the same GPU.
Inference latency is measured at batch sizes 1 and 64 and reported per window. Training uses batch size 64 and each model's training objective.
For the baselines, training-step time is estimated from the difference between four-epoch and one-epoch runtimes divided by the additional optimization steps. TENET is timed over warmed-up optimization steps.
Parameter storage and peak allocated training memory are reported in decimal MB.
Operation counts use \texttt{fvcore}, counting one multiply--accumulate as one operation, with analytical recurrent-layer counts added for operators not traced by the profiler.

\begin{table}[H]
\caption{Cross-model computational cost at $N=L=100$. Inference latency is per window, and training latency is per step. $B$ denotes batch size.}
\label{tab:cross_model_cost}
\centering
\setlength{\tabcolsep}{3pt}
\renewcommand{\arraystretch}{1.1}
\begin{scriptsize}
\begin{tabular}{c|cc|cc|cc}
\toprule
\textbf{Method} & \makecell{\textbf{Parameters}\\\textbf{(MB)}} & \makecell{\textbf{Operations}\\\textbf{(G)}}
& \makecell{\textbf{Inference}\\$B=1$ \textbf{(ms)}} & \makecell{\textbf{Inference}\\$B=64$ \textbf{(ms)}}
& \makecell{\textbf{Training}\\$B=64$ \textbf{(ms)}} & \makecell{\textbf{Training Peak}\\$B=64$ \textbf{(MB)}} \\
\midrule
THOC & 0.53 & 0.033 & 1.67 & 0.0267 & 3.63 & 439.07 \\
OmniAnomaly & 1.42 & 0.035 & 0.31 & 0.0059 & 2.05 & 224.62 \\
Anomaly Transformer & 19.82 & 0.526 & 0.85 & 0.0420 & 16.51 & 3,092.85 \\
CATCH & 4.39 & 0.665 & 1.69 & 0.4029 & 85.31 & 9,441.99 \\
SARAD & 38.45 & 0.986 & 0.44 & 0.0581 & 12.56 & 1,134.95 \\
DCdetector & 3.71 & 1.848 & 1.35 & 0.2530 & 53.93 & 8,183.42 \\
CrossAD & 10.58 & 4.133 & 0.78 & 0.5267 & 132.62 & 14,654.57 \\
\midrule
\textbf{TENET} & 700.49 & 3.151 & 33.18 & 1.2604 & 294.09 & 8,761.62 \\
\bottomrule
\end{tabular}
\end{scriptsize}
\end{table}

TENET has the largest parameter storage and higher measured latency in this comparison, reflecting the cost of its separate temporal encoders.
Its peak training memory is lower than that of CATCH and CrossAD, but higher than that of the other listed baselines.
Together with the detection results, this comparison makes the performance--cost trade-off explicit rather than suggesting that TENET is uniformly more efficient.

\newpage
\section{Representation Geometry Diagnostics}
\label{appendix:training_stability}

This appendix defines the diagnostics used in Figure~\ref{fig:collapse_combined}.
All diagnostics are computed from the pooled system state $\mathbf{c}=f_{\Theta}(\mathbf{X})\in\mathbb{R}^{h}$.
The covariance-based diagnostics use normal training windows to inspect the learned state geometry, while the radial and directional probes evaluate how much anomaly-discriminative information remains in different components of the state.

\paragraph{Covariance Spectrum.}
Let $\mathbf{C}=[\mathbf{c}_1,\dots,\mathbf{c}_B]^\top$ be a batch of pooled states and let $\bar{\mathbf{c}}=B^{-1}\sum_{i=1}^{B}\mathbf{c}_i$.
We compute the empirical covariance
\[
\Sigma
=
\frac{1}{B-1}
\sum_{i=1}^{B}
(\mathbf{c}_i-\bar{\mathbf{c}})
(\mathbf{c}_i-\bar{\mathbf{c}})^\top .
\]
Let $\lambda_1\ge\lambda_2\ge\cdots\ge\lambda_h\ge0$ be the eigenvalues of $\Sigma$.
This spectrum describes how the representation variance is distributed across latent directions.
A spectrum concentrated in a few large eigenvalues indicates that the state vectors occupy a low-dimensional subspace, whereas a flatter spectrum indicates a more distributed representation.

\paragraph{Effective Rank.}
Effective rank summarizes the dimensionality of the covariance spectrum~\cite{EffectiveRank-2007}.
With normalized eigenvalues $p_k=\lambda_k/\sum_{j=1}^{h}\lambda_j$, it is defined as
\[
\mathrm{eRank}(\Sigma)
=
\exp\!\left(
-\sum_{k=1}^{h} p_k \log p_k
\right).
\]
Higher effective rank means that variance is spread across more latent directions.
In Figure~\ref{fig:collapse_combined}(a), a higher value therefore indicates a less low-rank, less degenerate state geometry during training.

\paragraph{Cumulative Explained Variance Ratio.}
The cumulative explained variance ratio (EVR) shows how quickly the covariance spectrum concentrates.
For the top $m$ components, we compute
\[
\mathrm{EVR}(m)
=
\frac{\sum_{k=1}^{m}\lambda_k}
{\sum_{j=1}^{h}\lambda_j},
\qquad
m=1,\dots,h.
\]
If $\mathrm{EVR}(m)$ rises quickly, then a small number of principal directions accounts for most of the state variance.
If it rises slowly, the representation uses a broader portion of the state space.
Figure~\ref{fig:collapse_combined}(b) reports this curve after training.

\paragraph{Directional Probe.}
To test whether the state direction contains anomaly-relevant information, we construct a direction-only score.
For each state, define the unit direction
\[
\mathbf{u}_i
=
\frac{\mathbf{c}_i}{\|\mathbf{c}_i\|_2+\epsilon}.
\]
Here $\epsilon$ is a small constant used only for numerical stability.
We compute the mean normal direction from training states,
\[
\bar{\mathbf{u}}
=
\frac{\frac{1}{B}\sum_{i=1}^{B}\mathbf{u}_i}
{\left\|\frac{1}{B}\sum_{i=1}^{B}\mathbf{u}_i\right\|_2+\epsilon},
\]
and score a test state by cosine distance to this direction:
\[
s_{\mathrm{dir}}(\mathbf{c})
=
1
-
\frac{\mathbf{c}^{\top}\bar{\mathbf{u}}}
{\|\mathbf{c}\|_2+\epsilon}.
\]
This score removes radial magnitude and keeps only angular deviation from the normal direction.
Figure~\ref{fig:collapse_combined}(c) reports V-PR using this directional score.

\paragraph{Radial Probe.}
The norm-only score isolates radial information:
\[
s_{\mathrm{norm}}(\mathbf{c})
=
\|\mathbf{c}\|_2 .
\]
This score ignores angular information and uses only the distance of the state from the origin under the fixed standard normal reference.
Figure~\ref{fig:collapse_combined}(d) reports V-PR using this radial score.

\newpage
\section{Training and Test-Time Perturbation Sensitivity}
\label{appendix:contam}

\subsection{Training Contamination}

We evaluate whether TENET remains stable when perturbed windows are introduced into the normal training set.
At each optimization step, a random fraction of training windows, with ratio in $\{0,1,3,5,10,20\}\%$, is replaced by a perturbed version of the same window.
In this experiment, evaluation uses the original test split without added perturbations, and all other hyperparameters follow the main experimental protocol.

\begin{figure}[H]
\centering
\begin{tikzpicture}
\begin{groupplot}[
    group style={group size=4 by 1, horizontal sep=2em},
    width=0.29\linewidth,
    height=0.22\linewidth,
    ymajorgrids=true,
    grid style={dotted},
    x tick label style={font=\scriptsize},
    yticklabel style={font=\scriptsize},
    xlabel style={font=\scriptsize},
    ylabel style={font=\scriptsize},
    title style={font=\scriptsize, yshift=-6pt},
    every axis plot/.append style={thick, mark size=1.5pt},
    symbolic x coords={0,1,3,5,10,20},
    xtick=data,
    enlarge x limits=0.08,
]

\nextgroupplot[
    ymin=70, ymax=78,
    ylabel={V-PR (\%)},
    title={(a) PSM},
]
\addplot[name path=upper, draw=none, forget plot]
  coordinates {(0,75.94) (1,76.11) (3,76.10) (5,76.13) (10,76.12) (20,75.91)};
\addplot[name path=lower, draw=none, forget plot]
  coordinates {(0,72.46) (1,73.09) (3,73.50) (5,73.63) (10,73.32) (20,72.83)};
\addplot[fill=orange!25, forget plot] fill between[of=upper and lower];
\addplot+[draw=orange!80!black, mark=square*, mark options={fill=orange!80!black}]
  coordinates {(0,74.20) (1,74.60) (3,74.80) (5,74.88) (10,74.72) (20,74.37)};

\nextgroupplot[
    ymin=70, ymax=78,
    ylabel={},
    title={(b) SMD-7},
]
\addplot[name path=upper, draw=none, forget plot]
  coordinates {(0,75.73) (1,75.43) (3,75.54) (5,75.53) (10,75.47) (20,73.63)};
\addplot[name path=lower, draw=none, forget plot]
  coordinates {(0,72.31) (1,72.85) (3,72.74) (5,72.71) (10,72.07) (20,71.45)};
\addplot[fill=orange!25, forget plot] fill between[of=upper and lower];
\addplot+[draw=orange!80!black, mark=square*, mark options={fill=orange!80!black}]
  coordinates {(0,74.02) (1,74.14) (3,74.14) (5,74.12) (10,73.77) (20,72.54)};

\nextgroupplot[
    ymin=75, ymax=82,
    ylabel={},
    title={(c) SWaT},
]
\addplot[name path=upper, draw=none, forget plot]
  coordinates {(0,80.47) (1,80.33) (3,79.96) (5,79.93) (10,79.84) (20,79.21)};
\addplot[name path=lower, draw=none, forget plot]
  coordinates {(0,78.25) (1,78.29) (3,77.98) (5,77.71) (10,77.54) (20,76.69)};
\addplot[fill=orange!25, forget plot] fill between[of=upper and lower];
\addplot+[draw=orange!80!black, mark=square*, mark options={fill=orange!80!black}]
  coordinates {(0,79.36) (1,79.31) (3,78.97) (5,78.82) (10,78.69) (20,77.95)};

\nextgroupplot[
    ymin=79, ymax=83,
    ylabel={},
    title={(d) SWAN},
]
\addplot[name path=upper, draw=none, forget plot]
  coordinates {(0,81.68) (1,82.68) (3,82.71) (5,82.74) (10,82.79) (20,82.86)};
\addplot[name path=lower, draw=none, forget plot]
  coordinates {(0,80.26) (1,80.88) (3,80.87) (5,80.86) (10,80.85) (20,80.80)};
\addplot[fill=orange!25, forget plot] fill between[of=upper and lower];
\addplot+[draw=orange!80!black, mark=square*, mark options={fill=orange!80!black}]
  coordinates {(0,80.97) (1,81.78) (3,81.79) (5,81.80) (10,81.82) (20,81.83)};

\end{groupplot}
\end{tikzpicture}
\caption{
Uniform training perturbation robustness.
Panels report V-PR (\%) versus perturbation rate (\%) across datasets, with mean and $\pm 1$ standard deviation over five seeds.
}
\label{fig:contam_uniform}
\end{figure}
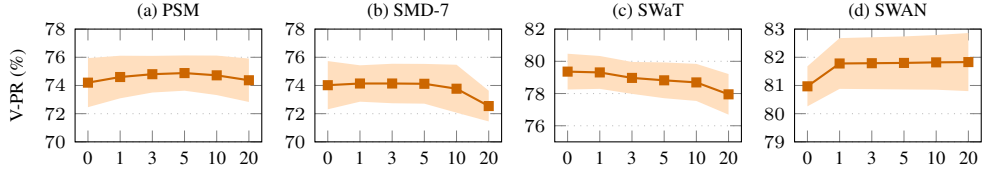

\paragraph{Zero-Mean Perturbations.}
We first consider per-point zero-mean perturbations using Gaussian noise $\mathcal{N}(0,\sigma^2)$ and variance-matched Uniform noise $\mathcal{U}(-\sigma\sqrt{3},\sigma\sqrt{3})$ with $\sigma=3.0$.
Figures~\ref{fig:contam} and~\ref{fig:contam_uniform} show that TENET remains relatively stable under these perturbations.
Performance changes modestly as the perturbation ratio increases, with drops remaining within roughly 1.6\%pt of the clean baseline even at the 20\% perturbation rate.
The Gaussian and uniform settings also produce similar trends.
These results suggest that the learned time-point representation is less sensitive to point-wise noise injected during training.

\begin{figure}[H]
\centering
\begin{tikzpicture}
\begin{groupplot}[
    group style={group size=4 by 1, horizontal sep=2em},
    width=0.29\linewidth,
    height=0.22\linewidth,
    ymajorgrids=true,
    grid style={dotted},
    x tick label style={font=\scriptsize},
    yticklabel style={font=\scriptsize},
    xlabel style={font=\scriptsize},
    ylabel style={font=\scriptsize},
    title style={font=\scriptsize, yshift=-6pt},
    every axis plot/.append style={thick, mark size=1.5pt},
    symbolic x coords={0,1,3,5,10,20},
    xtick=data,
    enlarge x limits=0.08,
]

\nextgroupplot[
    ymin=70, ymax=78,
    ylabel={V-PR (\%)},
    title={(a) PSM},
]
\addplot[name path=upper, draw=none, forget plot]
  coordinates {(0,75.94) (1,76.01) (3,75.94) (5,75.88) (10,75.72) (20,75.26)};
\addplot[name path=lower, draw=none, forget plot]
  coordinates {(0,72.46) (1,72.77) (3,73.18) (5,73.40) (10,73.18) (20,72.62)};
\addplot[fill=blue!15, forget plot] fill between[of=upper and lower];
\addplot+[draw=blue!70!black, mark=*, mark options={fill=blue!70!black}]
  coordinates {(0,74.20) (1,74.39) (3,74.56) (5,74.64) (10,74.45) (20,73.94)};

\nextgroupplot[
    ymin=70, ymax=78,
    ylabel={},
    title={(b) SMD-7},
]
\addplot[name path=upper, draw=none, forget plot]
  coordinates {(0,75.73) (1,75.45) (3,75.47) (5,75.28) (10,74.78) (20,74.18)};
\addplot[name path=lower, draw=none, forget plot]
  coordinates {(0,72.31) (1,72.75) (3,72.53) (5,72.32) (10,71.88) (20,71.68)};
\addplot[fill=blue!15, forget plot] fill between[of=upper and lower];
\addplot+[draw=blue!70!black, mark=*, mark options={fill=blue!70!black}]
  coordinates {(0,74.02) (1,74.10) (3,74.00) (5,73.80) (10,73.33) (20,72.93)};

\nextgroupplot[
    ymin=75, ymax=82,
    ylabel={},
    title={(c) SWaT},
]
\addplot[name path=upper, draw=none, forget plot]
  coordinates {(0,80.47) (1,80.48) (3,80.14) (5,79.94) (10,79.80) (20,79.12)};
\addplot[name path=lower, draw=none, forget plot]
  coordinates {(0,78.25) (1,78.24) (3,77.84) (5,77.78) (10,77.34) (20,76.44)};
\addplot[fill=blue!15, forget plot] fill between[of=upper and lower];
\addplot+[draw=blue!70!black, mark=*, mark options={fill=blue!70!black}]
  coordinates {(0,79.36) (1,79.36) (3,78.99) (5,78.86) (10,78.57) (20,77.78)};

\nextgroupplot[
    ymin=78, ymax=83,
    ylabel={},
    title={(d) SWAN},
]
\addplot[name path=upper, draw=none, forget plot]
  coordinates {(0,81.68) (1,81.68) (3,81.69) (5,81.66) (10,81.62) (20,81.61)};
\addplot[name path=lower, draw=none, forget plot]
  coordinates {(0,80.26) (1,80.20) (3,80.11) (5,80.02) (10,79.90) (20,79.73)};
\addplot[fill=blue!15, forget plot] fill between[of=upper and lower];
\addplot+[draw=blue!70!black, mark=*, mark options={fill=blue!70!black}]
  coordinates {(0,80.97) (1,80.94) (3,80.90) (5,80.84) (10,80.76) (20,80.67)};

\end{groupplot}
\end{tikzpicture}
\caption{
Gaussian training perturbation robustness.
Panels report V-PR (\%) versus perturbation rate (\%) across datasets, with mean and $\pm 1$ standard deviation over five seeds.
}
\label{fig:contam}
\end{figure}
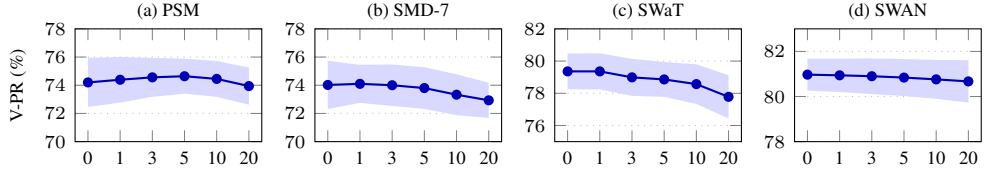

\begin{figure}[H]
\centering
\begin{tikzpicture}
\begin{groupplot}[
    group style={group size=4 by 1, horizontal sep=2em},
    width=0.29\linewidth,
    height=0.22\linewidth,
    ymajorgrids=true,
    grid style={dotted},
    x tick label style={font=\scriptsize},
    yticklabel style={font=\scriptsize},
    xlabel style={font=\scriptsize},
    ylabel style={font=\scriptsize},
    title style={font=\scriptsize, yshift=-6pt},
    every axis plot/.append style={thick, mark size=1.5pt},
    symbolic x coords={0,1,3,5,10,20},
    xtick=data,
    enlarge x limits=0.08,
]

\nextgroupplot[
    ymin=70, ymax=78,
    ylabel={V-PR (\%)},
    title={(a) PSM},
]
\addplot[name path=upper, draw=none, forget plot]
  coordinates {(0,75.94) (1,76.20) (3,76.33) (5,76.40) (10,76.01) (20,74.71)};
\addplot[name path=lower, draw=none, forget plot]
  coordinates {(0,72.46) (1,72.84) (3,72.91) (5,72.90) (10,72.27) (20,70.99)};
\addplot[fill=red!15, forget plot] fill between[of=upper and lower];
\addplot+[draw=red!80!black, mark=triangle*, mark options={fill=red!80!black}]
  coordinates {(0,74.20) (1,74.52) (3,74.62) (5,74.65) (10,74.14) (20,72.85)};

\nextgroupplot[
    ymin=68, ymax=76,
    ylabel={},
    title={(b) SMD-7},
]
\addplot[name path=upper, draw=none, forget plot]
  coordinates {(0,75.73) (1,74.78) (3,74.09) (5,73.76) (10,72.20) (20,70.65)};
\addplot[name path=lower, draw=none, forget plot]
  coordinates {(0,72.31) (1,72.12) (3,71.57) (5,70.78) (10,69.60) (20,68.31)};
\addplot[fill=red!15, forget plot] fill between[of=upper and lower];
\addplot+[draw=red!80!black, mark=triangle*, mark options={fill=red!80!black}]
  coordinates {(0,74.02) (1,73.45) (3,72.83) (5,72.27) (10,70.90) (20,69.48)};

\nextgroupplot[
    ymin=72, ymax=81,
    ylabel={},
    title={(c) SWaT},
]
\addplot[name path=upper, draw=none, forget plot]
  coordinates {(0,80.47) (1,79.02) (3,78.06) (5,77.52) (10,76.74) (20,76.23)};
\addplot[name path=lower, draw=none, forget plot]
  coordinates {(0,78.25) (1,76.30) (3,75.60) (5,75.20) (10,74.42) (20,72.73)};
\addplot[fill=red!15, forget plot] fill between[of=upper and lower];
\addplot+[draw=red!80!black, mark=triangle*, mark options={fill=red!80!black}]
  coordinates {(0,79.36) (1,77.66) (3,76.83) (5,76.36) (10,75.58) (20,74.48)};

\nextgroupplot[
    ymin=79, ymax=83,
    ylabel={},
    title={(d) SWAN},
]
\addplot[name path=upper, draw=none, forget plot]
  coordinates {(0,81.68) (1,82.65) (3,82.62) (5,82.58) (10,82.47) (20,82.22)};
\addplot[name path=lower, draw=none, forget plot]
  coordinates {(0,80.26) (1,80.81) (3,80.70) (5,80.60) (10,80.39) (20,79.98)};
\addplot[fill=red!15, forget plot] fill between[of=upper and lower];
\addplot+[draw=red!80!black, mark=triangle*, mark options={fill=red!80!black}]
  coordinates {(0,80.97) (1,81.73) (3,81.66) (5,81.59) (10,81.43) (20,81.10)};

\end{groupplot}
\end{tikzpicture}
\caption{
Constant-offset training perturbation robustness.
Panels report V-PR (\%) versus perturbation rate (\%) across datasets, with mean and $\pm 1$ standard deviation over five seeds.
}
\label{fig:contam_offset}
\end{figure}
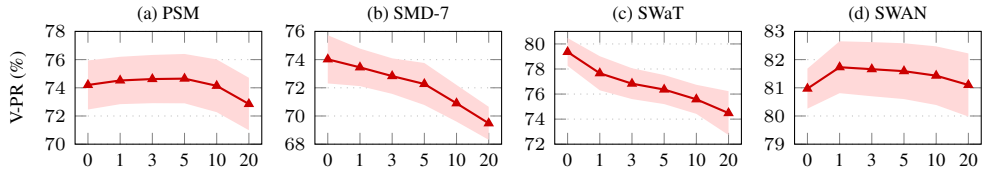

\paragraph{Coherent Offset Perturbations.}
We also test a more structured perturbation in which each selected window receives a per-variable constant offset drawn from $\mathcal{N}(0,\sigma^2)$.
Unlike point-wise zero-mean noise, this perturbation shifts an entire variable trajectory within a window and is therefore closer to calibration drift or a constant sensor offset.
Figure~\ref{fig:contam_offset} shows that offset perturbations are more challenging, especially on SMD and SWaT, where V-PR drops by up to roughly 4.9\%pt at the 20\% perturbation rate.
SWAN remains comparatively stable under the same perturbation.
Even in this more challenging setting, TENET retains useful detection performance, suggesting that the learned current-point representation remains stable under moderate training-set shifts.

\subsection{Current-Point Perturbations at Inference}
\label{appendix:current-point-noise}

Because the current observation determines the temporal query, perturbing it can change the retrieved context.
To measure this sensitivity, we add independent zero-mean Gaussian noise with standard deviation $\sigma$ to each variable of the current observation $\mathbf{x}_t$ in every test window on PSM.
Noise is added after standardization, while the preceding observations and original evaluation labels remain unchanged.
The model is not retrained, so this experiment treats the added perturbation as nuisance noise rather than relabeling it as an anomaly.
Figure~\ref{fig:current_point_noise} reports results over five seeds.

\definecolor{tenetBlue}{RGB}{70,130,180}

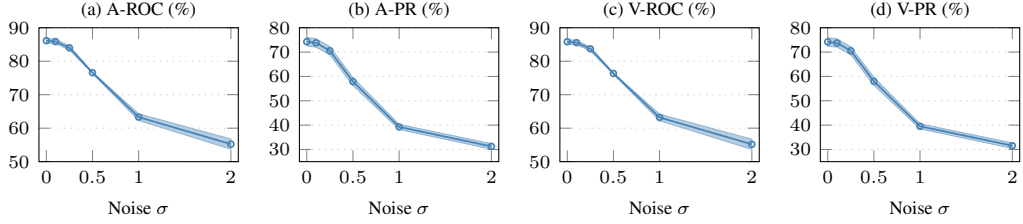
\begin{figure}[H]
\centering
\begin{tikzpicture}

\begin{groupplot}[
    group style={group size=4 by 1, horizontal sep=2.3em},
    width=0.302\linewidth,
    height=0.24\linewidth,
    xmin=0, xmax=2,
    enlarge x limits=0.04,
    xtick={0,0.5,1,2},
    xlabel={Noise $\sigma$},
    ymajorgrids=true,
    grid style={dotted},
    x tick label style={font=\scriptsize},
    yticklabel style={font=\scriptsize},
    xlabel style={font=\scriptsize},
    title style={font=\scriptsize, yshift=-6pt},
    every axis plot/.append style={thick, mark size=1.5pt},
]

\nextgroupplot[title={(a) A-ROC (\%)}, ymin=50, ymax=90, ytick={50,60,70,80,90}]
\addplot[name path=arocupper, draw=none, forget plot]
    table[col sep=comma, x=sigma, y expr=\thisrow{aroc}+\thisrow{aroc_std}]{appendices/current_point_noise.csv};
\addplot[name path=aroclower, draw=none, forget plot]
    table[col sep=comma, x=sigma, y expr=\thisrow{aroc}-\thisrow{aroc_std}]{appendices/current_point_noise.csv};
\addplot[fill=tenetBlue!40, draw=tenetBlue!60, line width=0.2pt, forget plot]
    fill between[of=arocupper and aroclower];
\addplot[draw=tenetBlue, line width=0.65pt, mark=o, mark size=1.1pt, mark options={fill=none}]
    table[col sep=comma, x=sigma, y=aroc]{appendices/current_point_noise.csv};

\nextgroupplot[title={(b) A-PR (\%)}, ymin=25, ymax=80, ytick={30,40,50,60,70,80}]
\addplot[name path=aprupper, draw=none, forget plot]
    table[col sep=comma, x=sigma, y expr=\thisrow{apr}+\thisrow{apr_std}]{appendices/current_point_noise.csv};
\addplot[name path=aprlower, draw=none, forget plot]
    table[col sep=comma, x=sigma, y expr=\thisrow{apr}-\thisrow{apr_std}]{appendices/current_point_noise.csv};
\addplot[fill=tenetBlue!40, draw=tenetBlue!60, line width=0.2pt, forget plot]
    fill between[of=aprupper and aprlower];
\addplot[draw=tenetBlue, line width=0.65pt, mark=o, mark size=1.1pt, mark options={fill=none}]
    table[col sep=comma, x=sigma, y=apr]{appendices/current_point_noise.csv};

\nextgroupplot[title={(c) V-ROC (\%)}, ymin=50, ymax=90, ytick={50,60,70,80,90}]
\addplot[name path=vrocupper, draw=none, forget plot]
    table[col sep=comma, x=sigma, y expr=\thisrow{vroc}+\thisrow{vroc_std}]{appendices/current_point_noise.csv};
\addplot[name path=vroclower, draw=none, forget plot]
    table[col sep=comma, x=sigma, y expr=\thisrow{vroc}-\thisrow{vroc_std}]{appendices/current_point_noise.csv};
\addplot[fill=tenetBlue!40, draw=tenetBlue!60, line width=0.2pt, forget plot]
    fill between[of=vrocupper and vroclower];
\addplot[draw=tenetBlue, line width=0.65pt, mark=o, mark size=1.1pt, mark options={fill=none}]
    table[col sep=comma, x=sigma, y=vroc]{appendices/current_point_noise.csv};

\nextgroupplot[title={(d) V-PR (\%)}, ymin=25, ymax=80, ytick={30,40,50,60,70,80}]
\addplot[name path=vprupper, draw=none, forget plot]
    table[col sep=comma, x=sigma, y expr=\thisrow{vpr}+\thisrow{vpr_std}]{appendices/current_point_noise.csv};
\addplot[name path=vprlower, draw=none, forget plot]
    table[col sep=comma, x=sigma, y expr=\thisrow{vpr}-\thisrow{vpr_std}]{appendices/current_point_noise.csv};
\addplot[fill=tenetBlue!40, draw=tenetBlue!60, line width=0.2pt, forget plot]
    fill between[of=vprupper and vprlower];
\addplot[draw=tenetBlue, line width=0.65pt, mark=o, mark size=1.1pt, mark options={fill=none}]
    table[col sep=comma, x=sigma, y=vpr]{appendices/current_point_noise.csv};

\end{groupplot}
\end{tikzpicture}
\caption{Sensitivity of TENET to test-time current-point noise on PSM. Panels report each metric (\%) versus noise standard deviation $\sigma$. Lines and markers show the mean over five seeds, and shaded bands indicate $\pm 1$ standard deviation.}
\label{fig:current_point_noise}
\end{figure}

Performance changes little at $\sigma=0.1$, but decreases as the noise grows, with substantial degradation at higher noise levels.
These results show tolerance to small current-point perturbations, not invariance to arbitrary corruption.

\newpage
\section{Broader Impacts}
\label{appendix:broader_impacts}

This work studies unsupervised anomaly detection for multivariate time series, with potential applications in industrial monitoring, server operations, scientific instruments, and cyber-physical systems.
Improved detection can help operators identify abnormal system behavior earlier, reduce downtime, and support more reliable maintenance decisions.
By showing that one-class TSAD can be a strong alternative when paired with appropriate representation construction, this work broadens the methodological choices available for both practical applications and future TSAD research.
The representation-focused design of TENET may also be useful as a general state encoder for downstream monitoring tasks beyond binary anomaly scoring.

At the same time, anomaly detectors should be deployed with care.
False positives can increase operator burden, while false negatives can delay intervention during real faults.
TENET provides anomaly scores rather than automated corrective actions, so practical deployments should combine the model with domain-specific thresholds, human review, and post-deployment monitoring.
Because real systems often experience distribution shift, calibration should be revisited after changes in sensors, operating policies, or system configuration.

The method is evaluated on public benchmark datasets and synthetic robustness settings.
These benchmarks are useful for controlled comparison, but they do not cover every deployment scenario.

\section{Limitations and Future Work}
\label{appendix:limitations}

TENET is designed to isolate the effect of current-point system state construction under a simple one-class scoring rule.
This design makes the empirical analysis cleaner, but it also leaves several limitations.

First, TENET uses inter-variable self-attention to compose variable-level states.
This allows flexible cross-variable modeling, but its cost grows with the number of variables.
The scalability study shows that the current implementation remains practical for the tested regimes, yet larger industrial systems may require sparse, hierarchical, or graph-constrained inter-variable layers.
Future work should therefore develop scalable inter-variable composition modules that preserve current-point-centric state construction while reducing the cost of multivariate composition in large sensor systems.
For long-window training, smaller micro-batches with gradient accumulation and activation checkpointing can reduce memory requirements.
Compressing long histories through downsampling or shorter key--value sequences is another possible direction for future work, but its effect on detection performance remains to be evaluated.

Second, a deep theoretical explanation of why TENET works well remains open.
The experiments with multiple one-class objectives show that TENET's advantage is not tied to a particular scoring rule and provide strong evidence for the utility of its architecture.
A deep theoretical explanation of why TENET produces an effective anomaly-ranking structure for one-class MTSAD would be a valuable direction for future work.

\end{document}